\def\year{2021}\relax
\documentclass[letterpaper]{article} 
\usepackage{aaai21}  
\usepackage{times}  
\usepackage{helvet} 
\usepackage{courier}  
\usepackage[hyphens]{url}  
\usepackage{graphicx} 
\usepackage{natbib}  
\usepackage{subcaption}
\usepackage{caption} 
\usepackage{amsmath}
\usepackage{algorithm}
\usepackage{algpseudocode}

\usepackage{adjustbox}
\usepackage{array}
\newcolumntype{R}[2]{%
    >{\adjustbox{angle=#1,lap=\width-(#2)}\bgroup}%
    l%
    <{\egroup}%
}

\newcommand*\rot{\rotatebox{80}}

\title{Circles are like Ellipses, or Ellipses are like Circles? Measuring the Degree of Asymmetry of Static and Contextual Word Embeddings and the Implications to Representation Learning}
\author {
    Wei Zhang \textsuperscript{\rm 1},
    Murray Campbell \textsuperscript{\rm 1}, 
    Yang Yu\footnote{work done while with IBM} \textsuperscript{\rm 2}, 
    Sadhana Kumaravel \textsuperscript{\rm 1}
    \\
}
\affiliations {
    \textsuperscript{\rm 1} IBM Research AI; 
    \textsuperscript{\rm 2} Google \\
    zhangwei@us.ibm.com, mcam@us.ibm.com, yangyuai@google.com, sadhana.kumaravel1@ibm.com
}

\begin{document}

\maketitle

\begin{abstract}
Human judgments of word similarity have been a popular method of evaluating the quality of word embedding. But it fails to measure the geometry properties such as asymmetry. For example, it is more natural to say ``Ellipses are like Circles'' than ``Circles are like Ellipses''. Such asymmetry has been observed from a psychoanalysis test called word evocation experiment, where one word is used to recall another. Although useful, such experimental data have been significantly understudied for measuring embedding quality. In this paper, we use three well-known evocation datasets to gain insights into asymmetry encoding of embedding. We study both static embedding as well as contextual embedding, such as BERT. Evaluating asymmetry for BERT is generally hard due to the dynamic nature of embedding. Thus, we probe BERT's conditional probabilities (as a language model) using a large number of Wikipedia contexts to derive a theoretically justifiable Bayesian asymmetry score. The result shows that contextual embedding shows randomness than static embedding on similarity judgments while performing well on asymmetry judgment, which aligns with its strong performance on ``extrinsic evaluations'' such as text classification. The asymmetry judgment and the Bayesian approach provides a new perspective to evaluate contextual embedding on intrinsic evaluation, and its comparison to similarity evaluation concludes our work with a discussion on the current state and the future of representation learning.
\end{abstract}

\section{Introduction}
Popular static word representations such as word2vec~\citep{mikolov2013distributed} lie in Euclidean space and are evaluated against symmetric judgments. Such a measure does not expose the geometry of word relations, e.g., asymmetry. For example, ``ellipses are like circles'' is much more natural than ``circles are like ellipses''. An acceptable representation may exhibit such a property.

~\citet{tversky1977features} proposed a similarity measure that encodes asymmetry.
It assumes each word is a feature set, and asymmetry manifests when the common features of two words take different proportions in their respective feature sets, i.e., a difference of the likelihoods $P(a|b)$ and $P(b|a)$ for a word pair ($a$,$b$). In this regard, the degree of correlation between asymmetry obtained from humans and a word embedding may indicate the quality of the embedding of encoding features.

Word evocation experiment devised by neurologist Sigmund Freud around the 1910s was to obtain such word directional relationship, where a word called cue is shown to a participant who is asked to ``evoke'' another word called target freely. ~\footnote{A clip from A Dangerous Method describing Sigmund Freud's experiment \url{https://www.youtube.com/watch?v=lblzHkoNn3Q}} The experiment is usually conducted on many participants for many cue words. The data produced from the group of people exhibit a collective nature of word relatedness. The $P(a|b)$ and $P(b|a)$ can be obtained from such data to obtain an asymmetry ratio~\citep{griffiths2007topics} that resonates with the theory of~\citet{tversky1977features} and ~\citet{Resnik:1995:UIC:1625855.1625914}. 
Large scale evocation datasets had been created to study the psychological aspects of language. We are interested in three of them; the Edinburgh Association Thesaurus
~\citep{kiss1973associative}, Florida Association Norms
~\citep{nelson2004university} and Small World of Words
\cite{de2019small} 
Those three datasets have thousands of cue words each and all publicly available. We use them to derive the human asymmetry judgments and see how well embedding-derived asymmetry measure aligns with this data.

Evocation data was rarely explored in the Computational Linguistics community, except that~\citet{griffiths2007topics} derived from the Florida Association Norms an asymmetry ratio for a pair of words to measure the directionality of word relations in topic models, and~\citet{nematzadeh2017evaluating} used it for word embedding. In this paper, we conduct a larger scale study using three datasets, on both static embedding (word2vec)~\citep{mikolov2013distributed}, GloVe ~\citep{pennington2014glove}, fasttext~\citep{mikolov2018advances}) and contextual embedding such as BERT~\citep{devlin2018bert}. We hope the study could help us better understand the geometry of word representations and inspire us to improve text representation learning.

To obtain $P(a|b)$ for static embedding, we leverage vector space geometry with projection and soft-max similar to ~\citep{nematzadeh2017evaluating,Levy2014NeuralWE,arora-etal-2016-latent}; For contextual embedding such as BERT we can not use this method because the embedding varies by context. Thus, we use a Bayesian method to estimate word conditional distribution from thousands of contexts using BERT as a language model. In so doing, we can probe the word relatedness in the dynamic embedding space in a principled way.

Comparing an asymmetry measure to the popular cosine measure, we observe that similarity judgment fails to correctly measure BERT's lexical semantic space, while asymmetry judgment shows an intuitive correlation with human data. In the final part of this paper, we briefly discuss the result and what it means to representation learning. \textit{This paper makes the following contributions:} 
\begin{enumerate}
    \item An analysis of embedding asymmetry with evocation datasets and an asymmetry dataset to facilitate research;
    \item An unbiased Bayesian estimation of word pair relatedness for contextual embedding, and a justifiable comparison of static and contextual embeddings on lexical semantics using asymmetry.
\end{enumerate}

\section{Related Work}

Word embedding can be evaluated by either the  symmetric ``intrinsic'' evaluation such as word pair similarity/relatedness \cite{agirre-etal-2009-study,hill2015simlex} and analogy \cite{mikolov2013distributed} or the ``extrinsic'' one of observing the performance on tasks such as text classification \cite{joulin2016fasttext}, machine reading comprehension \cite{rajpurkar-etal-2016-squad} or language understanding benchmarks \cite{wang2018glue,wang2019superglue}. 

On utterance-level, probing contextual embeddings was conducted mainly on BERT \cite{devlin2018bert}, suggesting its strength in encoding syntactic information rather than semantics \cite{hewitt2019structural,reif2019visualizing,tenney2019you,tenney2019bert,mickus2019mean}, which is counter-intuitive given contextual representation's superior performance on external evaluation.  

On the lexical level, it is yet unknown if previous observation also holds. Moreover, lexical-semantic evaluation is non-trivial for contextual embedding due to its dynamic nature. Nevertheless, some recent approach still tries to either extract a static embedding from BERT using PCA \cite{Ethayarajh2019HowCA,coenen2019visualizing}, or use context embeddings as is \cite{mickus2019mean} for lexical-semantic evaluation. Some instead use sentence templates~\citep{petroni2019language,bouraoui2019inducing} or directly analyze contextual embedding for its different types of information other than lexical ones \cite{brunner2019validity,clark2019what,coenen2019visualizing,jawahar2019does}. So far, a theoretically justifiable method still is missing to mitigate the bias introduced in the assumption of above analysis methods. Thus, how contextual and static embedding compares on lexical semantics is still open. 

An asymmetry ratio was used to study the characteristics of static embedding \cite{nematzadeh2017evaluating} and topic models \cite{griffiths2007topics}, and the study of asymmetry of contextual embedding is still lacking. Given the current status of research, it is natural to ask the following \textit{research questions}:
\begin{itemize}
    \item 
    \noindent{\textbf{RQ 1.}} Which evocation dataset is the best candidate to obtain asymmetry ground-truth? And \text{\textbf{RQ 1.1}:} Does the asymmetry score from data align with intuition? \text{\textbf{RQ 1.2}:} Are evocation datasets correlated for the first place? 
    \item
    \noindent{\textbf{RQ 2.}} How do static and context embeddings compare on asymmetry judgment? Furthermore, how do different context embeddings compare? Do larger models perform better?
    \item
    \noindent{\textbf{RQ 3.}} What are the main factors to estimating $P(b|a)$ with context embedding, and what are their effects?
    \item
    \noindent{\textbf{RQ 4.}} Does asymmetry judgment and similarity judgment agree on embedding quality? What does it imply?
\end{itemize}

We first establish the asymmetry measurement in Section \ref{sec:asy_measure} and methods to estimate them from an evocation data or an embedding in Section \ref{sec:pba} followed by empirical results to answer those questions in Section \ref{sec:exp}.

\section{The Asymmetry Measure} 
\label{sec:asy_measure}
\subsubsection{Log Asymmetry Ratio (LAR) of a word pair.} To measure asymmetry of a word pair $(a;b)$ in some set of word pairs $\mathcal{S}$, we define two conditional likelihoods, $P_\mathcal{E}(b|a)$ and $P_\mathcal{E}(a|b)$, which can be obtained from a resource $\mathcal{E}$, either an evocation dataset or embedding. Under Tversky's ~\shortcite{tversky1977features} assumption, if $b$ relates to more concepts than $a$ does, $P(b|a)$ will be greater than $P(a|b)$ resulting in asymmetry. A ratio $P_\mathcal{E}(b|a)/P_\mathcal{E}(a|b)$ \cite{griffiths2007topics} can be used to quantify such asymmetry. We further take the logarithm to obtain a \textit{log asymmetry ratio} (LAR) so that the degree of asymmetry can naturally align with the sign of LAR (a ratio close to 0 suggests symmetry, negative/positive otherwise). Formally, the LAR of $(a;b)$ from resource $\mathcal{E}$ is 
\begin{equation}
\label{eq:lar}
    \mbox{LAR}_\mathcal{E}(a;b) = \log P_\mathcal{E}(b|a) - \log P_\mathcal{E}(a|b)
\end{equation}
LAR is key to all the following metrics.
\subsubsection{Aggregated LAR (ALAR) of a pair set.}
We care about the aggregated LAR on a word pair set $\mathcal{S}$, the expectation
$    \mbox{ALAR}_\mathcal{E}(\mathcal{S}) = \mathbf{E}_{(a;b)\in \mathcal{S}} [\mbox{LAR}_\mathcal{E}(a;b)]
$, to quantify the overall asymmetry on $\mathcal{S}$ for $\mathcal{E}$. However, it is not sensible to evaluate ALAR on any set of $\mathcal{S}$, for two reasons:  1) because $\text{LAR(a;b)} = - \text{LAR}(b;a)$, randomly ordered word pairs will produce random ALAR values; 2) pairs of different relation types may have very different LAR signs to cancel each other out if aggregated. For example, ``$a$ is a part of $b$'' suggest LAR$(a;b)>0$ 
, and ``$a$ has a $b$'' for LAR$(a;b)< 0$. Thus, we evaluate ALAR on $\mathcal{S}(r)$, the relation-specific subset, as 
\begin{equation}
    \label{eq:alar}
    \mbox{ALAR}_\mathcal{E}(\mathcal{S}(r)) = \mathbf{E}_{(a,b)\in { \mathcal{S}(r)} } [\mbox{LAR}_\mathcal{E}(a;b)]
\end{equation}
where the order of $(a,b)$ is determined by $(a,r,b)\in \mbox{KG}$, the ConceptNet \cite{speer2017conceptnet}. Note that when $\mathcal{E}$ is an evocation data, we can qualitatively examine if $\mbox{ALAR}_\mathcal{E}(\mathcal{S}(r))$ aligns with with human intuition as one (\textbf{RQ 1.2}) of the two metrics to determine the candidacy of an evocation data as ground truth; When $\mathcal{E}$ is any embedding, it measures the asymmetry of embedding space.

\subsubsection{Correlation on Asymmetry (CAM) of Resources.}
By using LAR defined in Eq.~\ref{eq:lar}, for a resource $\mathcal{E}$ and a word pair set $\mathcal{S}$ we can define a word-pair-to-LAR map 
\begin{equation}
\label{eq:m}
\mathcal{M}(\mathcal{E},\mathcal{S}) = \{(a;b):\text{LAR}_\mathcal{E}(a;b) | (a;b) \in \mathcal{S}\}
\end{equation}
and the Spearman Rank Correlation on asymmetry measure (CAM) between two resources $\mathcal{E}_i$ and $\mathcal{E}_j$ can be defined as
\begin{equation}
    \text{CAM}(\mathcal{S}, \mathcal{E}_i, \mathcal{E}_j)=\text{Spearman}(\mathcal{M}(\mathcal{E}_i,\mathcal{S}), \mathcal{M}(\mathcal{E}_j,\mathcal{S}))
    \label{eq:cam}
\end{equation} 
There are two important settings: 
\begin{enumerate}
    \item $\mathcal{E}_i$ is an evocation data and $\mathcal{E}_j$ is an embedding
    \item $\mathcal{E}_i$ and $\mathcal{E}_j$ are two different evocation datasets 
\end{enumerate}
Setting one is to evaluate embedding using evocation data as the ground truth. Setting two is to measure the correlation between any pair of evocation data, which is the second metric (\textbf{RQ 1.1}) to validate the candidacy of an evocation dataset as asymmetry ground-truth: several studies indicate that the human scores consistently have very high correlations with each other \cite{miller1991contextual,Resnik:1995:UIC:1625855.1625914}. Thus it is reasonable to hypothesize that useful evocation data should correlate stronger in general with other evocation data. We will discuss it more in experiments.
\begin{table*}
\centering
\begin{tabular}{@{}c|cc|cc|cc|ccc@{}}
   & \multicolumn{2}{c}{EAT} & \multicolumn{2}{c}{FA}  & \multicolumn{2}{c}{SWOW} & \multicolumn{3}{c}{Spearman's Correlation (\textbf{CAM})}\\

$r$ & count & \textbf{ALAR} & count & \textbf{ALAR} & count & \textbf{ALAR} & EAT-FA & SW-FA & SW-EAT \\
\hline

 relatedTo         & 	8296      & 	4.50     &	5067    & 	0.89    & 	34061     & 	4.83       & 	0.59 & 	0.68 & 	0.64 \\
antonym           & 	1755      & 	1.27     &	1516    & 	0.38    & 	3075      & 	0.01       & 	0.43 & 	0.58 & 	0.51 \\
synonym           & 	673       & 	-15.80   &	385     & 	-17.93  & 	2590      & 	-15.85     & 	0.49 & 	0.65 & 	0.59 \\
isA               & 	379       & 	43.56    &	342     & 	31.59   & 	1213       & 	47.77      & 	0.64 & 	0.75 & 	0.59 \\
atLocation        & 	455       & 	17.48    &	356     & 	9.59    & 	1348       & 	16.02      & 	0.61 & 	0.71 & 	0.64 \\
distinctFrom      & 	297       & 	-2.38    &	250     & 	0.01    & 	593       & 	-1.07      & 	0.32 & 	0.57 & 	0.43 \\
\end{tabular}
\caption{Pair count, \textbf{ALAR}($S(r)$) and the Spearman Correlation on Asymmetry Measure (\textbf{CAM}) between datasets. See Appendix for a complete list. P-value$<0.00001$ for all \textbf{CAM} results}

\label{tab:1}
\end{table*}

Next, we introduce how to obtain $P(b|a)$ for calculating LAR, ALAR, and CAM from different resources. 

\section{Method} \label{sec:pba}

\subsection{$P_D(b|a)$ from Evocation Data}
For evocation data $D$, $P_D(b|a)$ means when $a$ is cue, how likely $b$ can be evoked \cite{kiss1973associative}. During the experiment, one has to go through a thought process to come up with $b$. This process is different from  humans writing or speaking with natural languages, because the word associations are free from the basic demands of communication in natural language, making it an ideal tool to study internal representations of word meaning and language in general \cite{de2019small}. Unlike similarity judgment where a rating (usually 1 to 10) is used to indicate the degree of similarity of $(a,b)$, evocation data does not immediately produce such a rating, but a binary ``judgment'' indicating if $b$ is a response of $a$ or not. Yet, evocation data is collected from a group of participants~\citep{de2019small,nelson2004university}, and we could derive a count-based indicator to average the binary judgments to give a rating as is done for the similarity judgment averaging the scores of experts. The total number of responses usually normalizes such count, leading to a score called \textit{Forward Association Strength} (FSG)~\citep{nelson2004university}, a metric invented for psychology study. FSG score is essentially the $P_D(b|a)$,
\begin{equation}
    P_D(b|a) = \frac{ \text{Count}(b \mbox{ as a response} | a \mbox{ is cue}) } { \text{Count} (a \mbox{ is cue})}
\end{equation}

It is easy to confuse such evocation counts with the counts derived from texts. Again, the counts in evocation data are the aggregation of judgments \cite{de2019small} rather than co-occurrence from language usage which is subject to the demands of communication.


\subsection{$P_B(b|a)$ from Contextual Embedding} 
It is easy to obtain $P(b|a)$ for static embedding by exploring geometric properties such as vector projection. But estimating it with contextual embedding is generally hard due to the embedding's dynamic nature invalidating the projection approach. 
Thus, to evaluate $P(b|a)$ within a contextual embedding space in an unbiased manner yet admitting its dynamic nature, we first find the contexts that $a$ and $b$ co-occur, and then use one word to predict the likelihood of another, admitting the existence of the context. Finally, to remove the bias introduced by context, we average the likelihood over many contexts to obtain an un-biased estimate of $P(b|a)$. This idea can be understood in a Bayesian perspective: we introduce a random variable $\textbf{c}$ to denote a paragraph as context from corpus $C$, say, Wikipedia. Then we obtain the expectation of $P_B(b|a)$ over $\mathbf{c}$, using $B$, say BERT, as a language model as $P_B(b|a)= \mathbf{E}_{P(\mathbf{c}|a), \mathbf{c} \in C} [P_B(b|a,\mathbf{c})]$. The formation can be simplified: for $\textbf{c}$ that does not contain $a$, $P(\textbf{c}|a)=0$, and for $\textbf{c}$ that does not contain $b$, $P_B(b|a,\textbf{c})=0$. Finally,
\begin{equation} \label{eq:bert}
    P_B(b|a)=\mathbf{E}_{P(\mathbf{c}|a), \mathbf{c} \in C(\{a,b\})} [P_B(b|\mathbf{c})]
\end{equation}
where $C(x)$ indicates all the contexts in $C$ that includes $x$, being either a single word or a word pair. Note that $P_B(b|a,\mathbf{c})=P_B(b|\mathbf{c})$ if $a\in \mathbf{c}$, leading to Eq.~\ref{eq:bert}. $P(\textbf{c}|a)$ is estimated as $1/|C(a)|$ and $P_B(b|\textbf{c})$ is estimated by masking $b$ from the whole paragraph $\textbf{c}$ and then getting the probability of $b$ from the Soft-max output of a pre-trained BERT-like Masked language model $B$. When there are N words of b in $\mathbf{c}$, we only mask the one that is being predicted and repeat the prediction for each b. We append ``[CLS]'' to the beginning of a paragraph and ``[SEP]''after each sentence. Word $a$ can also appear $k>1$ times in $\mathbf{c}$ and we regard $\mathbf{c}$ as $k$ contexts for $C(a)$. 

Using BERT as a language model, we can make an unbiased estimation of context-free word relatedness, if the contexts are sufficient and the distribution is not biased. But, like all unbiased estimators, Eq.~\ref{eq:bert} may suffer from high variance due to the complexity of context $\textbf{c}$. We identify two factors of context that may relate to estimation quality (\textbf{RQ 3}): the number of contexts and the distance between words as a bias of context, which we discuss in the experiments.

\subsection{$P_E(b|a)$ from Static Embedding} 
To answer \textbf{RQ 4}, we also calculate conditionals for static embedding. For word2vec or GloVe, each word can be regarded as a bag of features ~\citep{tversky1977features} and $P(b|a)$ can be obtained using a ``normalized intersection'' of the feature sets for $a$ and $b$ which corresponds to geometric projection in continuous embedding vector space:
\begin{equation}
\label{eq:logit}
    proj(b|a) = \frac{emb(b) \cdot emb(a)}{||emb(a)||}
\end{equation}
And we normalize them with Soft-max function to obtain $P_E$ as
\begin{equation}
\label{eq:softmax}
    P_E(b|a) = \frac{exp(proj(b|a))}{\sum_x exp(proj(x|a)}
\end{equation} 
where the range of $x$ is the range of the evocation dataset.
If we compare Eq.~\ref{eq:softmax} and ~\ref{eq:logit} to the dot-product (the numerator in Eq.~\ref{eq:logit}) that is used for similarity measurement~\citep{Levy2014NeuralWE, nematzadeh2017evaluating, arora-etal-2016-latent}, we can see dot-product only evaluates how much overlap two embedding vectors have in common regardless of its proportion in the entire meaning representation of $a$ or $b$. In other words, it says ``ellipses'' are similar to ``circles''. But it fails to capture if there is more to ``circles'' that is different than ``ellipses''.

\subsubsection{An Asymmetry-Inspired Embedding and Evaluation.}
Unlike static embedding that embeds all words into a single space, a word can have two embeddings. In word2vec learning \cite{mikolov2013distributed}, it corresponds to the word and context embedding learned jointly by the dot-product objective. Intuitively, such a dot-product could encode word relatedness of different relation types more easily than with a single embedding space. To verify this, we use the weight matrix in word2vec skip-gram as context embedding similar to~\citep{torabi-asr-etal-2018-querying} together with the word embedding to calculate $P_E(b|a)$. We denote it as \textbf{cxt}. With \textbf{cxt}, the asymmetry can be explicitly encoded: to obtain $P(b|a)$, we use word embedding for $a$ and context embedding for $b$, and then apply Eq.~\ref{eq:logit} and ~\ref{eq:softmax}.

\section{Result and Analysis}
\label{sec:exp}
To sum up, we first obtain $P(b|a)$ (Section \ref{sec:pba}) from evocation data, static and context embedding respectively, and then use them to calculate asymmetry measure LAR, ALAR and CAM (Section \ref{sec:asy_measure}). Below we start to answer the plethora of research questions with empirical results.

\subsection{\textbf{RQ 1}: Which evocation data is better to obtain asymmetry ground truth?} 
We answer it by examining two sub-questions: an evocation data's correlation to human intuition (\textbf{RQ 1.1}) and its correlation with other evocation data (\textbf{RQ 1.2}).

\subsubsection{RQ 1.1: Correlation to Human Intuition}
To see if an evocation data conforms to intuition, we examine the ALAR for each relation type $r$ separately, which requires grouping word pairs $\mathcal{S}$ to obtain $\mathcal{S}(r)$.

Unfortunately, evocation data does not come with relation annotations. Thus we use ConceptNet~\citep{speer2017conceptnet} to automatically annotate word relations. Specifically, we obtain $S(r)=\{(a,b)\}$ where $(a,b)$ is connected by $r$ (we treat all relations directional) where $a$ as head and $b$ is tail. If $(a,b)$ has multiple relations \{$r_i$\}, we add the pair to each $S(r_i)$. Pairs not found are annotated with the ConceptNet's \textit{relatedTo} relation. 
Finally we calculate the ALAR$_\mathcal{E}(\mathcal{S}(r)$) using Eq.~\ref{eq:alar}
for each relation $r$.

Table \ref{tab:1} shows a short list of relation types and their ALAR($\mathcal{S}(r)$). We can see that \textit{isA}, \textit{partOf}, \textit{atLocation}, \textit{hasContext} exhibit polarized ALAR; whereas \textit{antonym} and \textit{distinctFrom} are comparably neutral. These observations agree with intuition in general, except for \textit{synonym} and \textit{similarTo}, (e.g., ``ellipses'' and ``circles'') which ALARs show mild asymmetry. This may have exposed the uncertainty and bias of humans on the KG annotation of similarity judgments, especially when 
multiple relations can be used to label a pair, e.g. ``circle'' and ``ellipse'' can be annotated with either ``isA'' or ``similarTo''. However, such bias does not affect the correctness of asymmetry evaluation because the Spearman correlation of two resources is correctly defined no matter which set of pairs is used. 
The relation-specific ALAR is for qualitatively understanding the data. However, this interesting phenomenon may worth future studies.

\subsubsection{RQ 1.2: Correlation to each other}
\label{sec:correlation}
The observation that good human data have high correlations with each other \cite{miller1991contextual, Resnik:1995:UIC:1625855.1625914} provides us a principle to understand the quality of the three evocation datasets by examining how they correlate. Our tool is CAM of Eq. \ref{eq:cam} defined on the common set of word pairs, the intersection $\mathcal{S}(r)=\mathcal{S}_{\text{EAT}}(r) \cap \mathcal{S}_{\text{FA}}(r) \cap \mathcal{S}_{\text{SWOW}}(r)$ where each set on RHS is the collected pairs for $r$ in a dataset. The number of common pairs for $r$ is about 90\% of the smallest dataset for each $r$ in general. Then, the CAM is obtained by Eq.~\ref{eq:m} and \ref{eq:cam}, e.g. CAM$_{\mathcal{S}(r)}$(SWOW, FA). The calculated CAM is shown in Table~\ref{tab:1}, and a longer list is in Table \ref{tab:full_data_lar_cam} in Appendix. In general, SWOW and FA show stronger ties, probably because they are more recent and closer in time; EAT correlates less due to language drift. 
\begin{table*}
\centering
\begin{tabular}{@{}l|cccc|cccc|cccc@{}}
  & \multicolumn{4}{c}{\textbf{EAT} (12K pairs)} & \multicolumn{4}{c}{\textbf{FA} (8K pairs) }  & \multicolumn{4}{c}{\textbf{SWOW} (30K pairs)}\\

  & w2v/cxt &glv & fxt & bert/bertl & w2v/cxt &glv & fxt & bert/bertl& w2v/cxt &glv & fxt & bert/bertl\\

 \hline

relatedTo &.08/.25&.37&.20& .48/\textbf{.55} &.17/.30 &.41&.27&.44/\textbf{.50}&.06/.28&.42&.14& .43/\textbf{.50}\\

antonym&.05/.15 &.25&.07&.31/\textbf{.38} &.16/.23&.31&.21&.30/\textbf{.38}&.04/.15& .33&.09& .33/\textbf{.41}\\

synonym &-.21/.14&.43&-.03&.52/\textbf{.59} &.19/.37&\textbf{.44}&.40&.33/.41&.00/.29& .43 &.16& .38/\textbf{.47} \\

isA&.06/.33&.45&.27&.50/\textbf{.58} &.23/.41&.44&.37&.43/\textbf{.50}&.03/.34&.39&.16&.50/\textbf{.57} \\

atLocation&.08/.28&.44&.29&.45/\textbf{.52} &.22/.36&\textbf{.47}&.31&.33/.44&.08/.35&.47&.21&.44/\textbf{.52} \\

distinctFrom&-.12/-.20&.05&-.09&.17/\textbf{.28} &.11/.17&.35&.17&.34/\textbf{.42}&.03/.19& .40&.16&.38/\textbf{.49}\\
\hline
\textbf{SA} & .05/.22 & .34 & .16 & .45/\textbf{.52} & .17/.29 &.39 & .26 & .38/\textbf{.46} & .05/.27 & .41 & .15 & .42/\textbf{.49}\\
\hline
\textbf{SR} & -.02/.15 & .30 & .08 & .40/\textbf{.47} & .17/.27 & .36 & .25 & .35/\textbf{.43} & .02/.24 & .38 & .16 & .40/\textbf{.48} \\

\end{tabular}
\caption{Spearman Correlation on Asymmetry Measure (\textbf{CAM}) between embedding LAR and data LAR. Acronyms: w2v (word2vec), glv (GloVe), fxt (fasttext), bert (\texttt{BERT-base}), bertl (\texttt{BERT-large}).
\textbf{SA} (Weight-Averaged Spearman, where weights are calculated as $|S(r)|/|S|$); \textbf{SR} (\textbf{SA} excluding \textit{relateTo} relation). P-value$<$0.0001 for BERT and GloVe in general. Using Eq.~\ref{eq:common_pairs}, where $V$ is the intersection of above embeddings, we collect 12K pairs for EAT data, 8K for FA, and 30K for SWOW.
}

\label{tab:2}
\end{table*}

\subsubsection{Answering RQ 1: Which data to use?} From Table \ref{tab:1}, we favor SWOW because 1) in the study of \textbf{RQ 1.1} we see SWOW aligns with human intuition as well as, if not better than, the other two datasets, e.g., it made almost symmetric ALAR estimation on pair-abundant relations such as  \textit{antonym}; 2) According to the answer to \textbf{RQ 1.2}, in general, SWOW correlates to all other datasets the best, e.g., on the most pair-abundant relation \textit{relatedTo}, SWOW has the top two CAM scores, 0.68 and 0.64 to other datasets; 3) it is the largest and the most recent dataset. Thus we mainly use SWOW for later discussions.

\subsection{\textbf{RQ 2}: Asymmetry of Embedding} 
\subsubsection{Setup.} We compare embeddings using CAM in Eq.~\ref{eq:cam} and set $\mathcal{E}_i$ to an embedding $E$ and $\mathcal{E}_j$ to evocation data $D$ using LAR obtained according to Section \ref{sec:pba}. For context embeddings, we leverage the masked language models obtained from Huggingface Toolkit ~\citep{Wolf2019HuggingFacesTS} (See Appendix Table~\ref{tab:modellist} for a full list of models), and for static embeddings we both obtain pre-trained embeddings (for GloVe and fasttext) and train embeddings ourselves (w2v and cxt) using Wikipedia corpus (October 2019 dump, details are in Appendix). An issue that hampers fair comparison of embeddings is their vocabularies differ. To create a common set of word pairs, we take the intersection $V$ of the vocabularies of all embeddings that are to be compared and the evocation dataset, and for Eq.~\ref{eq:cam} we obtain $\mathcal{S}(r)$ as
\begin{equation}
\label{eq:common_pairs}
 \{(a,b) | (a,b) \in D \wedge r(a,b)\in \text{KG} \wedge a\in V \wedge b\in V\} 
\end{equation}
where KG is ConceptNet. which means any word in the set of pairs has to be in-vocabulary for any embedding. 

We have two settings: 1) comparing static and contextual embeddings (BERT as a representative), wherein applying Eq. \ref{eq:common_pairs} leads to 12K, 8K, and 30K pairs on EAT, FA, and SWOW dataset in Table \ref{tab:2}.
 2) comparing contextual embeddings, which leads to 7.3K SWOW pairs with asymmetry scores that will be made public.

\subsubsection{Comparing Static and Contextual embedding} In Table \ref{tab:2}, we compare BERT (base and large) with static embeddings on three different evocation datasets with \textbf{CAM} (Relation-specific ALAR can provide us a qualitative understanding how embedding performs on each relation in general but it does not affect the correctness of CAM). GLoVe is the most competitive among static embeddings because it takes into account the ratio $P(x|a_1)/P(x|a_2)$ that may help learn $P(b|a)$, which can lead to better accuracy on some relations. BERT, especially \texttt{BERT-large}, has a stronger correlation with the three datasets than any other static embedding, which aligns with the empirical evidence from external evaluation benchmarks such as SQUAD \cite{rajpurkar-etal-2016-squad} and GLUE \cite{wang2018glue}. It may be the first time we can show context embedding outperforms static embedding on intrinsic evaluation. Moreover, by comparing \textbf{CAM} on a per-relation basis, we see BERT performs competitively on LAR-polarized, asymmetric relations such as ``relatedTo'', ``isA'' and ``atLocation'', while not so much on symmetric ones. Also, the context embedding (cxt) consistently outperforms word2vec on almost all relation types. Combining these observations, we think that the dot-product of two embedding spaces can encode rich information than a single embedding space can. BERT does it with a key-query-value self-attention mechanism, being one reason for it to perform well on the asymmetry judgment. Also, it is not surprising that \texttt{BERT-large} outperforms \texttt{BERT-base}, suggesting larger models can indeed help better ``memorize'' word semantics, which we also show for other models soon later.

\subsubsection{What about LAR Directions?} 
\begin{figure}
  \includegraphics[width=\linewidth]{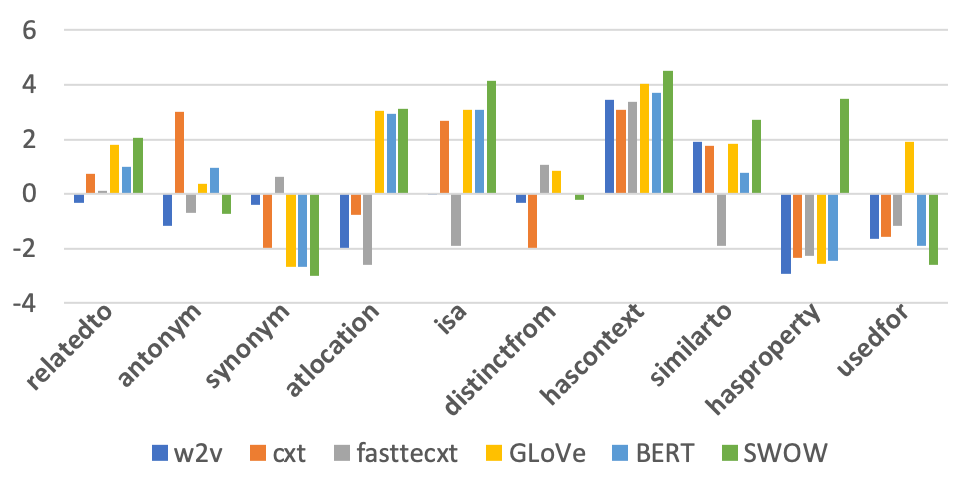}
  \caption{LAR Prediction Comparison. Same setting as Table~\ref{tab:2}. BERT refers to \texttt{BERT-base}}
  \label{fig:lar_dir}
\end{figure}

An embedding could have a high correlation (CAM) with data but totally wrong on asymmetry directions (LAR). Thus in Fig.~\ref{fig:lar_dir}, we compare embeddings' ALAR to the ALAR of data (SWOW). We took $log$ over the ALAR($r$) while retaining the sign to smooth out the numbers. \texttt{BERT-base} produces small ALAR values, which we scale by x1000 before $log$ to make the figure easy to read. \texttt{BERT-base} and GLoVe are two strong embeddings that show better directional correlation with SWOW. Note that word pairs under \textit{hasContext} aligns with SWOW data generally well, but words with relations \textit{hasProperty}, \textit{capableOf}, \textit{hasA} is hard for all text embeddings. These findings may suggest a non-negligible gap (Spearman's Correlation for BERT-SWOW and GloVe-SWOW has P-value$<0.0001$) between text-based word embeddings and the human-generated evocation data, regardless of how embeddings are trained. It may be either because texts do not entail relations of those pairs or because relations are too hard for current embedding techniques to discover, which requires further investigation. 

\begin{table}
\centering
\begin{tabular}{lr|llllll}
&
\rot{\# Params (M)} &
\rot{relatedTo} &
\rot{antonym} &
\rot{synonym} &
\rot{isA} &
\rot{atLocation} &
\rot{distinctFrom}
\\
\hline
\texttt{\textbf{BERT}} && \\
\texttt{-base}    &     110  & .33  &  .23 & .32 & .36 & .44 & .23    \\ 
\texttt{-large}        & 340     & .41 &.26 &  .38 & .45 & .49     &  .28    \\
\texttt{\textbf{roBERTa}} &&\\
\texttt{-base}        & 125 &  .41 &  .27 & .38 & .45 & .48 & .27    \\ 
\texttt{-large}        & 355 & .42 &  .27 & .38 & .46 & .49 & .29   \\ 
\texttt{\textbf{ALBERT}} &&\\
\texttt{-base}        & 11 &  .36 & .24 & .37 &.42 & .41 & .25    \\ 
\texttt{-large}        & 17 & .38  & .25 & .38 & .44 & .42 & .24 \\ 
\texttt{-xl}        & 58 &  .39 &  .26 & .37 & .45 & .43 & .24    \\ 
\texttt{-xxl}        & 223 &  .39  &  .25 & .37 & .43    & .44 & .26 \\ 
\texttt{\textbf{ELECTRA}} &&\\
\texttt{-base}        & 110 &  .39 &  .24 & .38 & .45 & .44 & .24    \\ 
\texttt{-large}        & 335 & .36  &  .26 & .35 & .34 & .42 & .29    \\ 

\end{tabular}

\caption{Spearman Correlation on Asymmetry Measure (CAM) between SWOW and contextual embedding for 7.3K SWOW pairs obtained with Eq.~\ref{eq:common_pairs}. Counts: 5409, 910, 217, 211, 262,  206 from left to right. }
\label{tab:compare_cxt_emb}
\end{table}

\subsubsection{Comparing Contextual Embeddings.}
By applying Eq.~\ref{eq:common_pairs}, we use all candidate contextual embeddings' vocabulary to obtain $V$ and use SWOW as $D$, resulting in 7.3K pairs in total, for which in Table~\ref{tab:compare_cxt_emb} we show the comparison of embeddings on CAM. In general, we see that larger models indeed show a higher correlation with SWOW data, which suggests models with larger capacity can help encode lexical semantics better in general. For example, CAM of BERT \cite{devlin2018bert}, roBERTa \cite{liu2019roberta} and ALBERT \cite{lan2019albert} grow with the number of parameters, yet ELECTRA \cite{clark2020electra} does not show the same trend. One reason may be that ELECTRA uses generated text for training, which may result in data drift that is exacerbated by using larger models. While ELECTRA and ALBERT have shown their advantage over BERT in many external evaluation benchmarks such as GLUE, SUPERGLUE, and SQuAD \cite{rajpurkar-etal-2016-squad,wang2018glue,wang2019superglue}, they do not improve significantly over asymmetry judgment. It is reasonable to doubt if the improvements come from better model tuning or better semantic representation, and asymmetry judgment may shed some light on the answer to this question. Also, RoBERTa outperform BERT may be because RoBERTa improves BERT's optimization, and in turn it confirms that optimization matter on semantic encoding.

\begin{figure*}
  \includegraphics[width=\linewidth]{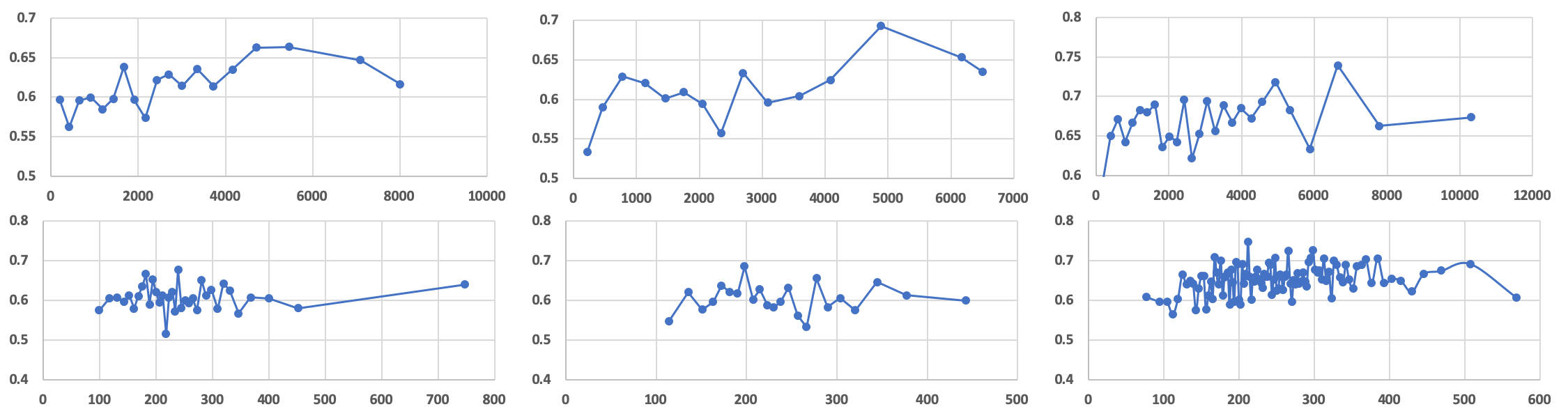}
  \caption{BERT directional accuracy (y-axis) v.s. context frequency (x-axis for upper 3 figures) and average character distance (x-axis lower 3 figures) for EAT (left), FA (middle) and SWOW (right)}
  \label{fig:da_stat}
\end{figure*}

\subsection{RQ 3: Two Factors of the Bayesian Estimation}
Since we evaluate $P(b|a)$ with contexts, the quality and distribution of it matter the most for the LAR, ALAR, and CAM. Comparing \texttt{BERT-base} and SWOW as an example, we study two quantities: 1) The number of contexts of a word pair because more contexts suggest the more accurate the estimation may be; and 2) the distance of word pairs in context, because the closer two words are, the more likely they relate. For 1), we group word pairs into bins of size 200 by the number of contexts collected for each pair and use average LAR directional accuracy (Appendix~\ref{app:dir_acc}) in each bin as a tool to study the factors' impact. The three upper figures in Figure~\ref{fig:da_stat} suggest a mild trend where pairs with more contexts have higher direction accuracy, which discontinues beyond 5000 contexts. We hypothesis that the discontinuation is due to the pairs grouped into $>$5000 bin may contain ``systematical bias'', such as topics, where a word depends more on the topic than the other word, which pushes asymmetry prediction towards random. Techniques of diversifying the context may help alleviate the problem, an extensive study too complicated for this paper. For 2), we group word pairs by character distance into size-200 bins. The three bottom ones in Figure~\ref{fig:da_stat} show word distance correlates weakly to direction accuracy due to BERT's ability to model long-distance word dependency.
\begin{table*}
\centering

\begin{tabular}{rcccc|rrrrrrrrrrrr}
 & w2v & cxt & glv & fxt & brt & brtl & rbt& rbtl &abt & abtl & abtxl & abtxxl & elt& eltl& KEl1 & KEl12\\
 \hline
MEN & .74 & .76 & .74 & \textbf{.79} & \textbf{.26} & .07 & .11 & .11 & -.04 & -.16 & .00 & .01 & .07 & .08 & .20 & .08\\
SIMLEX & .33 & 34 &.37 & \textbf{.45} & \textbf{.36} & .16& .18 & .34 & .16 & .18 &.19 & .19 &.12 & .12 & .32 & .23  \\
WS353 & .70 & .71 & .64 & \textbf{.73} &.28 & .28 &.26 & .09 &.06 &.02 & .01 & -.01 & .17 & -.03 & \textbf{.39}&.33 \\
--SIM & .76 & .76 & .70 & \textbf{.81} &\textbf{.12} & .08 & .07 & -.02 & -.05 & -.08 & -.01 & .16 & .01 & -.14 & - & - \\
--REL & .64 & .67 & .60 & \textbf{.68}& .45 & .48 & .47 & \textbf{.49} & .23 & .12 &.11 & .26 & .34 & .09 & - & -  \\
\end{tabular}
\caption{Spearman Correlation between model scores and oracle word similarity datasets MEN \cite{bruni2014multimodal}, SIMLEX999 (SIMLEX) \cite{hill2015simlex}, WordSim353 \cite{finkelstein2002placing} similar (SIM) and relatedness (REL) subsets. KEl1 and KEl2 are two results listed in \cite{Ethayarajh2019HowCA} which extract static embeddings from  BERT-base model. l1 and l12 are principle components from first and last layer BERT embedding therein.}
\label{tab:3}
\end{table*}

\subsection{RQ 4: Similarity v.s. Asymmetry Judgment}
In comparison to asymmetry judgment, we would like to see if similarity judgment can say otherwise about embeddings. We compare all embeddings on popular symmetric similarity/relatedness datasets. We take the dot-product score of two word embeddings for static embeddings and calculate the Spearman's correlation on the scores. For contextual embeddings, we use the geometric mean of $P(a|b)$ and $P(b|a)$ as similarity score (see Appendix for justification) to be comparable to dot-product. Table~\ref{tab:3} shows that although this approach helps BERT performs better on 2 out of 3 datasets than PCA on the contextual embedding approach \cite{Ethayarajh2019HowCA}, the result on other contextual embeddings looks extremely arbitrary compared to static embeddings. Similarity judgment, in general, fails to uncover contextual embedding's ability on lexical semantics: it focuses on similarity rather than the difference that BERT seems to be good at, which can also be supported by contextual embeddings being superior on WS353 REL than SIM. 
Similarity judgment tells us that contextual embedding does not correctly encode semantic features and static embeddings, but it can beat them reasonably well on asymmetry judgment, suggesting otherwise. Are they conflicting with each other? Let us look into it now.

\section{Discussion and Conclusion}

The rise of Transformers has aroused much speculation on how the model works on a wide variety of NLP tasks. One lesson we learned is that learning from large corpora how to match contexts is very important, and many tasks require the ability. From the intrinsic evaluation perspective, the asymmetry judgment and similarity judgment also support this. BERT can encode rich features that help relatedness modeling, but it fails frustratingly on similarity judgments that suggest otherwise.

We should not take this contradiction of similarity and asymmetry judgment on BERT slightly. If correct, our analysis shows BERT can not encode the meaning of words as well as static embedding can, but it learns contextual matching so well that it supports the modeling of $P(b|a)$ to exhibit a correct asymmetry ratio. Does it make sense at all? It is not clear if BERT learns word meaning well because similarity judgment does not provide supporting evidence. It probably is hard, if not impossible, to extract a stable word meaning representation out of the rest of the information that the dynamic embedding can encode for context matching. Even if we can, the evaluation may not be sound since they probably are correlated in the BERT's decision making. 

But, do we even care about if Transformers encode meanings? Is it OK to encode an adequate amount and let context matching do the rest? On the one hand, feeding meanings to Transformers in the form of external hand-crafted knowledge has not been as successful as we had hoped, yet the work is still going on under this philosophy ~\citep{liu2019k}; On the other hand, we continue to relentlessly pursue larger models such as the unbelievably colossal GPT-3~\cite{brown2020language} with 175-billion-parameters, and Table~\ref{tab:compare_cxt_emb} shows that naively scaling up model size does not guarantee significantly better word relatedness modeling. It may be high time that we stop and think how far we should chug along the path of Transformers with bells and whistles, and if it can lead us from 0.99 to 1. Learning representation by capturing the world's complexity through automatic discovery \footnote{http://www.incompleteideas.net/IncIdeas/BitterLesson.html} may still be the way we can follow. However, we may need either a very different family of models that encode meaning and context matching together harmoniously or a new objective very different from predicting a masked word or the next sentence. We do not know if BERT will look like it, or it will look like BERT. It must be out there waiting for us to discover.

\section{Future Work} There are still many questions left: How does lexical semantics (through asymmetry judgment) encoded in contextual embeddings change before and after transfer learning (task-specific fine-tuning) or multi-task learning? which can also guide how much to fine-tune or what task to learn from; 
2) Does multi-modal word representations help lexical semantics in general and demonstrate better  performance on both similarity and asymmetry judgment?

\section{Acknowledgement}
We thank anonymous reviewers for their feedbacks, colleagues for their interest and the support from our families during this special time to make this work happen.

\bibliography{aaai21.bib}

\clearpage

\appendix

\onecolumn

\section*{APPENDICES}

\begin{table*}[ht]

\centering

\begin{tabular}{@{}r|cc|cc|cc|ccc@{}}

  & \multicolumn{2}{c}{EAT} & \multicolumn{2}{c}{FA}  & \multicolumn{2}{c}{SWOW} & \multicolumn{3}{c}{Spearman Correlation}\\


& count & ratio & count & ratio & count & ratio & EAT-FA & SW-FA & SW-EAT \\
\hline

 relatedTo         & 	8296      & 	4.50     &	5067    & 	0.89    & 	34061     & 	4.83       & 	0.59 & 	0.68 & 	0.64 \\
antonym           & 	1755      & 	1.27     &	1516    & 	0.38    & 	3075      & 	0.01       & 	0.43 & 	0.58 & 	0.51 \\
synonym           & 	673       & 	-15.80   &	385     & 	-17.93  & 	2590      & 	-15.85     & 	0.49 & 	0.65 & 	0.59 \\
isA               & 	379       & 	43.56    &	342     & 	31.59   & 	1213       & 	47.77      & 	0.64 & 	0.75 & 	0.59 \\
atLocation        & 	455       & 	17.48    &	356     & 	9.59    & 	1348       & 	16.02      & 	0.61 & 	0.71 & 	0.64 \\
distinctFrom      & 	297       & 	-2.38    &	250     & 	0.01    & 	593       & 	-1.07      & 	0.32 & 	0.57 & 	0.43 \\
similarTo         & 	100       & 	-23.57   &	93      & 	18.89   & 	377       & 	11.31      & 	0.63 & 	0.69 & 	0.60 \\
hasContext        & 	74        & 	56.67    &	43      & 	61.10   & 	406       & 	64.76      & 	0.49 & 	0.42 & 	0.40 \\
hasProperty       & 	77        & 	40.63    &	69      & 	29.05   & 	295       & 	34.37      & 	0.50 & 	0.76 & 	0.53 \\
partOf            & 	64        & 	66.30    &	61      & 	56.69   & 	208       & 	60.46      & 	0.43 & 	0.68 & 	0.56 \\
usedFor           & 	40        & 	9.13     &	19      & 	-16.65  & 	265       & 	-25.93     & 	0.04 & 	0.46 & 	0.39 \\
capableOf         & 	64        & 	-31.80   &	47      & 	4.42    & 	137       & 	-14.54     & 	0.73 & 	0.72 & 	0.68 \\
hasA              & 	78        & 	-58.03   &	30      & 	-107.98 & 	132        & 	-77.63     & 	0.53 & 	0.75 & 	0.68 \\
hasPrerequisite   & 	31        & 	15.67    &	28      & 	-58.34  & 	133        & 	17.48      & 	0.71 & 	0.72 & 	0.69 \\
causes            & 	33        & 	34.03    &	13      & 	31.69   & 	121        & 	9.91       & 	0.25 & 	0.73 & 	0.70 \\
formOf            & 	46        & 	82.47    &	12      & 	42.84   & 	81        & 	72.67      & 	0.09 & 	0.39 & 	0.61 \\
derivedFrom       & 	16        & 	23.91    &	9       & 	137.57  & 	106        & 	30.75      & 	0.00 & 	-0.25 & 	0.59 \\
madeOf            & 	18        & 	27.65    &	15      & 	62.56   & 	65        & 	-2.09      & 	0.43 & 	0.67 & 	0.40 \\
hasSubevent       & 	21        & 	1.97     &	13      & 	-63.40  & 	70        & 	-38.48     & 	0.74 & 	0.78 & 	0.50 \\
motivatedByGoal   & 	15        & 	2.24     &	19      & 	-1.11   & 	45        & 	20.05      & 	0.58 & 	0.79 & 	0.55 \\
causesDesire      & 	20        & 	-62.26   &	20      & 	-31.22  & 	49        & 	-57.72     & 	0.85 & 	0.67 & 	0.80 \\
createdBy         & 	10        & 	-50.65   &	9       & 	-57.84  & 	29        & 	-106.94    & 	-0.25 & 	0.26 & 	0.13 \\
hasFirstSubevent  & 	10        & 	77.19    &	11      & 	0.91    & 	24        & 	-15.03     & 	0.14 & 	0.37 & 	0.22 \\
desires           & 	7         & 	-38.10   &	6       & 	-131.95 & 	23        & 	-46.69     & 	0.80 & 	0.66 & 	0.66 \\
hasLastSubevent   & 	5         & 	-15.54   &	6       & 	-3.65   & 	14        & 	-2.36      & 	0.50 & 	0.44 & 	0.70 \\
definedAs         & 	4         & 	25.90    &	3       & 	-36.94  & 	8         & 	11.04      & 	1.00 & 	0.87 & 	0.95 \\
notHasProperty    & 	2         & 	-43.41   &	2       & 	121.35  & 	4         & 	-10.13     & 	1.00 & 	1.00 & 	1.00 \\

\end{tabular}

\caption{Per-relation pair count, LAR, Spearman Correlation on LAR among datasets. Small World of Words is from the preprocessed SWOW-EN2018 Preprocessed data. We show the full data statistics in this Table, for each dataset, we filter the word pairs that only has one direction frequency. For example, $P(a|b)$ and $P(b|a)$ will be filtered if \textit{b} is not a target of \textit{a} or \textit{a} is not a target of \textit{b}. 
This guarantees that estimating missing probabilities does NOT play a part in pair evaluation. It is more likely that the missing pair not being there indicates the pair has close to zero probability in that direction than being accidentally missing from the dataset since the data is collected from the crowd. 
The result of this is a collection of ``clean pairs'' that has a higher correlation in general than overall statistics collected from all the pairs regardless of missing directional pairs. 
}

\label{tab:full_data_lar_cam}
\end{table*}

\begin{table*}[ht]
\centering

\begin{tabular}{@{}l|ccccc|ccccc|cccccc@{}}



  & \multicolumn{5}{c}{EAT (12K pairs)} & \multicolumn{5}{c}{FA (8.3K pairs)}  & \multicolumn{6}{c}{SWOW (35K pairs) }\\


  & w2v & cxt &glv & fxt & bertl & w2v & cxt &glv & fxt & bertl & w2v & cxt &glv & fxt & bertl\\

 \hline

relatedTo&.08&.25&.37&.20&\textbf{.48}&.17&.30&.41&.27&\textbf{.44}&.06&.28&.42&.14 & .43 & \textbf{.50}\\
antonym&.05&.15&.25&.07&\textbf{.31}&.16&.23&\textbf{.31}&.21&.30&.04&.15&.33 &.09&.33 & \textbf{.41}\\
synonym&-.21&.14&.43&-.03&\textbf{.52}&.19&.37&\textbf{.44}&.40&.33&.00&.29& .43 &.16&.38 & \textbf{.47}\\
isA&.06&.33&.45&.27&\textbf{.50}&.23&.41&\textbf{.44}&.37&.43&.03&.34&.39&.16&.44 & \textbf{.52}\\
atLocation&.08&.28&.44&.29&\textbf{.45}&.22&.36&\textbf{.47}&.31&.33&.08&.35&.47&.21&.50 & \textbf{.57}\\
DistinctFrom&-.12&-.20&.05&-.09&\textbf{.17}&.11&.17&\textbf{.35}&.17&.34&.03&.19& .40 &.16&.38 & \textbf{.49}\\
similarTo&-.18&.14&.18&.17&\textbf{.54}&.02&.37&\textbf{.45}&\textbf{.45}&.31&-.08&.19&.38&.27&.41 & \textbf{.56}\\
hasContext&.32&.30&\textbf{.49}&.22&.39&.07&.06&\textbf{.37}&.10&.29&.02&.24&.42&.29&.40 & \textbf{.54}\\
hasProperty&-.43&-.03&-.12&-.04&\textbf{.39}&-.28&.00&.26&-.00&\textbf{.39}&-.17&.12&.43&.09&.44 & \textbf{.50}\\
partOf&.18&.23&\textbf{.52}&.26&.46&.50&\textbf{.64}&.57&.51&.44&.43&.63& .65&.52&.58 &\textbf{.66}\\
capableOf&.45&.31&\textbf{.70}&.34&.47&.33&.30&\textbf{.58}&.26&.06&.13&.33&\textbf{.82}&.05&.48 & .61\\
hasPrereq.&.55&\textbf{.90}&.83&.52&.55&.37&.46&\textbf{.56}&.42&.38&-.46&.01&\textbf{.62}&-.46 &.46 & .56\\
hasA&-.41&-.37&-.04&.00&\textbf{.41}&.03&.26&.23&.13&\textbf{.63}&-.23&.12&-.19&.44&.47 & \textbf{.60}\\
derivedFrom&-.50&-1.00&-.50&-.50&\textbf{-.12}&-.56&-.67&-.56&-.05&\textbf{-.04}&.10&.20&.30&\textbf{.60}&.40 & .52\\
\hline
\textbf{SA} & .05 & .22 & .34 & .16 & \textbf{.45} & .17 & .29 & \textbf{.39} & .26 & .38 & .05 & .27 & .41 & .15 & .42 & \textbf{.49}\\
\hline
\textbf{SR} & -.02 & .15 & .30 & .08 & \textbf{.40} & .17 & .27 & \textbf{.36} & .25 & .35 & .02 & .24 & .38 & .16 & .40 & \textbf{.48}\\

\end{tabular}
\caption{Embedding's LAR Spearman Correlation with data. 
SA=Weight-Averaged Spearman; SR=SA without relateTo; hasPrereq.=hasPrerequisite. The table displays top relations that have relatively sufficient amount of pairs for evaluation (determined by P-value$<$0.001). The recall for each pair (the existence of the word in each representation's vocabulary) is reasonably high and in general, yet BERT-base (bert) has a smaller vocabulary 30K words. From the \textit{relatedTo} type, which has the most pairs to compare, we can be relatively confident to compare the LAR Spearman correlations among representations.
}

\label{tab:5}
\end{table*}

\begin{figure*}[ht]
  \includegraphics[width=\linewidth]{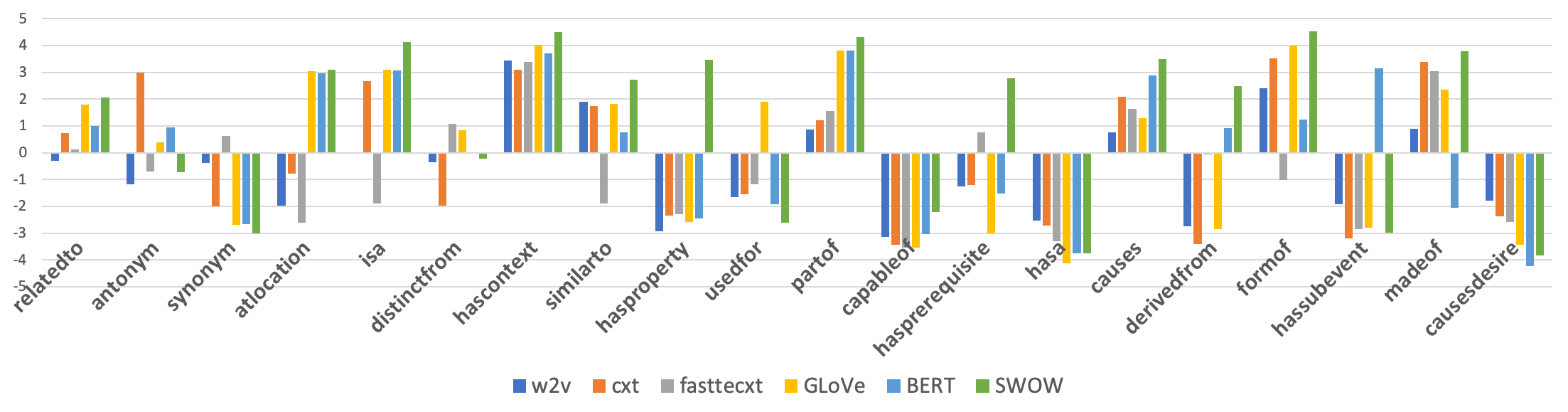}
  \caption{Comparison of LAR absolute values for a more detailed list of relations. The order of columns in graph is in the same order as legend (from left to right). We adjusted the ALAR of Embeddings to be roughly on the same scale by scaling BERT-base scores x1000.}
  \label{fig:abs_lar_detail}
\end{figure*}

\begin{figure*}[ht]
  \includegraphics[width=\linewidth]{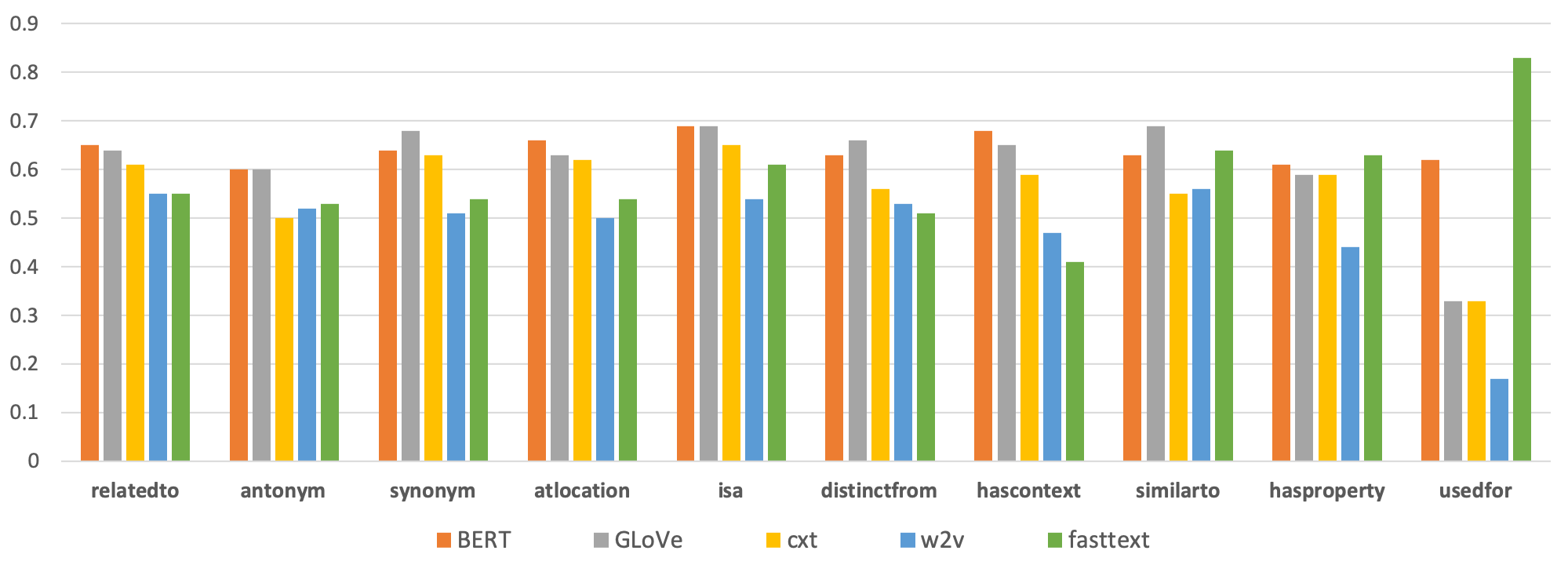}
  \caption{Direction Accuracy on SWOW comparing static embeddings and contextual embedding. The order in graph is the same as legend from left to right. We show the directional accuracy in this Figure. BERT refers to the BERT-base model ($\gamma=10$ in Eq. \ref{eq:diracc}). }
  \label{fig:dir_acc}
\end{figure*}

\section{Evocation Data}
The Edinburgh Association Thesaurus \footnote{http://vlado.fmf.uni-lj.si/pub/networks/data/dic/eat/Eat.htm}  (EAT)~\citep{kiss1973associative}, Florida Association Norms \footnote{http://w3.usf.edu/FreeAssociation/} (FA)~\citep{nelson2004university} and Small World of Words \footnote{https://smallworldofwords.org/en/project/research} (SWOW) \cite{de2019small} (the pre-processed 2018 data). 

\section{Extracting Context Details}
We first build an inverted index using October 2019 Wikipedia dump using wikiextractor \cite{Wikiextractor2015} as the tool for pulling pure text out of each wiki page and Whoosh \footnote{https://whoosh.readthedocs.io/en/latest/intro.html} to index the texts. Each document has been split into paragraphs as a context unit, and for querying a word pair, we use two words as two separate terms and use AND to join them. The retrieved contexts are used to obtain BERT LM predictions. For estimating $C(b)$ for a single word $b$, we use the single word as query and count all matched documents. A list of models used are in Table \ref{tab:modellist}, where the function used \texttt{XXFromMaskedLM} class where \texttt{XX} can be any of the contextual embedding name. Refer to \texttt{encoder.py} in code appendix for details.

\section{System Specifics and Running Time for Contextual Embeddings}
The indexing of Wikipedia using Whoosh toolkit took less than 12 hours. Using the largest SWOW data as an example, we obtain 100K (105581) word pairs, each word pair on average obtained roughly 1K contexts. Using one context embedding to annotate all word pairs to obtain both $P(b|a)$ and $P(a|b)$ took from 6 hours for \texttt{ALBERT-base} to longest 18 hours for \texttt{ALBERT-xxlarge}, using 350 nodes on a computing cluster where each node is equipped with Redhat Linux 7.0 with Intel(R) Xeon(R) CPU E5-2667 v2 @ 3.30GHz and  Nvidia K40 GPU. Running all contextual embedding models for SWOW took five days. We use Python 3.7 with Pytorch 1.2.0 as Huggingface backend. All models are run on GPU for MaskedLM inference. There is no randomness in any algorithm running process; thus, we ran each experiment once. 

\section{Static Embedding Details}
For \textbf{fasttext} we use the pre-trained 1 million 300-dimensional word vectors trained on Wikipedia 2017, UMBC webbase corpus and statmt.org news dataset (16B tokens) downloadable from \texttt{https://fasttext.cc}

For \textbf{GLoVe} embedding, we use Common Crawl pre-trained model (42B tokens, 1.9M vocab, uncased, 300d vectors, 1.75 GB download) \footnote{https://nlp.stanford.edu/projects/glove/}. 

For \textbf{word2vec}, we use original C code \footnote{https://github.com/tmikolov/word2vec/blob/master/word2vec.c}, and train it using Mikolov's \cite{mikolov2013distributed} original C code on October 2019 Wiki dump using nltk tokenizer \texttt{nltk.tokenize.regexp}. We set context window size 20 to increase contextuality and embedding size 300, a universal embedding size and set the rest as default. For \textbf{cxt} we extract the weight embedding out as the representation of context embedding.

\section{Word similarity/relatedness Measure Detail} We show a more comprehensive list in Table \ref{tab:6}. The datasets are MEN \cite{bruni2014multimodal}, SIMLEX999 \cite{hill2015simlex}, MT771, MT287, WS353 \cite{finkelstein2002placing} relation subset (REL) and similarity subset (SIM), YP130 \cite{yang2006verb}, RG65 \cite{rubenstein1965contextual}, which are popular sets used for word relatedness evaluation. 

We pick the datasets because they are all focused on word relatedness evaluation and relatively popular. We exclude the analogy datasets because they focus only on ``similarity'' relation, which scope is pretty limited. The datasets that show in Table \ref{tab:3} are either relatively large, have data comparable to KE19\cite{Ethayarajh2019HowCA}, or show the difference between similarity and relation to observe the embeddings' differences.

For static embeddings, the similarity measure is using Cosine similarity between word embeddings; for BERT, we use the geometric mean of the two directional probabilities in order to align with word embedding's cosine similarity. If we think of $P(a|b)$ and $P(b|a)$ are two projections, then cosine similarity is exactly the geometric mean of those two projection proportions which we can see below.

\section{Similarity measure of Contextual Embedding}
We can only derive $P(b|a)$ and $P(a|b)$, but not the $cosine(a,b)$. But,

\begin{equation}
\begin{split}
cosine(a,b) =& \frac{a \cdot b}{|a||b|} \\
=& \sqrt{\frac{a \cdot b  a \cdot b }{|a|^2|b|^2}} \\
=& \sqrt{\frac{a \cdot b }{|a|}  \frac{1}{|a|} \frac{ a \cdot b }{|b|}\frac{1}{|b|}} \\
=&\sqrt{P(b|a) P(a|b)}\sqrt{\frac{1}{|a||b|}}
\end{split}
\end{equation}

\begin{equation}
    sim(a,b) = \sqrt{\frac{P(a|b)P(b|a)}{|a||b|}}
\end{equation}

Here we assume $|a||b|$ as a constant. This approximation can not be justified easily. It is a much more complex discussion on how to ``de-bias'' embeddings using our Bayesian approach, so that a truly justified approach can be proposed to potentially improve this method.

\begin{equation}
sim(a,b) = \sqrt{P(a|b)P(b|a)}
\end{equation}

\begin{table*}[!]
\centering
\begin{tabular}{l|l}
\textbf{Model} & \textbf{Huggingface model name}  \\
\hline
\texttt{BERT-base}     & bert-base-uncased  \\
\texttt{BERT-large}     &  bert-large-uncased \\
\texttt{roBERTa-base}     &  roberta-base \\
\texttt{roBERTa-large}     & roberta-large \\
\texttt{ALBERT-base}     &  albert-base-v2 \\
\texttt{ALBERT-large}     &  albert-large-v2 \\
\texttt{ALBERT-xlarge}     &  albert-xlarge-v2\\
\texttt{ALBERT-xxlarge}     & albert-xxlarge-v2 \\
\texttt{ELECTRA-base}     & google/electra-base-discriminator \\
\texttt{ELECTRA-large}     & google/electra-large-discriminator \\

     & 
\end{tabular}
\caption{A list of contextual embeddings from Huggingface that are used in this paper. For all hyper-parameter settings we use model-accompanied default values. ELECTRA is trained as a discriminator/generator model where the discriminator (similar to BERT) is trained with synthetically generated data from generator instead of real sentences.}
\label{tab:modellist}
\end{table*}

\begin{table*}[!]
\centering
\begin{tabular}{r|rr|rr|rrrr|rr}
 & bert & bert-l & roberta & roberta-l &albert &albert-l &albert-xl & albert-xxl & electra & electra-l\\
 \hline

MEN & 25.87 & 7.93 & 11.06 & 7.57 & -3.70 & -16.49 &-0.34& 1.30 & 6.48& 7.50\\
SIMLEX & 36.17 & 16.12 & 18.18 & 33.67 & 16.47 & 17.9 & 19.33 & 19.11 &12.23& 11.27\\
MT771  & 24.48 & 17.91 & 18.69 & 18.67 & 16.54 & 19.92 &23.48 &14.78& 16.66 & 9.80\\
MT287  &33.71& -1.46 & -7.61 & 82.86 & 18.28 & 23.29 &30.95 & 23.60& 28.11 & 6.10\\
WS353  &27.51 &28.06 &26.41 & 8.57 & 6.29 & 2.08 &1.09 & 0.58& 16.98& -2.52\\
--SIM &12.24 & 8.83 & 7.15 & -1.67 & -5.43 & -8.24 &0.98 & 15.61& 0.88& -14.13\\
--REL &45.09 & 48.14 & 47.30 & 49.80 & 22.77 & 12.26 &11.56& 25.51& 33.83 & 9.25\\
\end{tabular}
\caption{Spearman Correlation between model scores and oracle word similarity datasets. KE19 \cite{Ethayarajh2019HowCA}, L1 and L12 are principle component from first and last layer BERT-base embedding; brt (BERT-base), brt-l (BERT-large), rbt (Roberta-base), rbt-l (Roberta-large), elt (ELECTRA-base), elt-l (ELECTRA-large), abt (Albert-base), abt-l (Albert-large), abt-xl (Albert-xlarge), abt-xxl (Albert-xxlarge). Each embedding is evaluated individually against datasets. If a word in a pair does not exist, it is excluded from the pair. for evaluation.}
\label{tab:6}
\end{table*}

\section{Directional Accuracy of Embedding LARs}
\label{app:dir_acc}
\begin{figure*}[!]
\centering

\begin{minipage}[b]{\textwidth}
  \includegraphics[width=\linewidth]{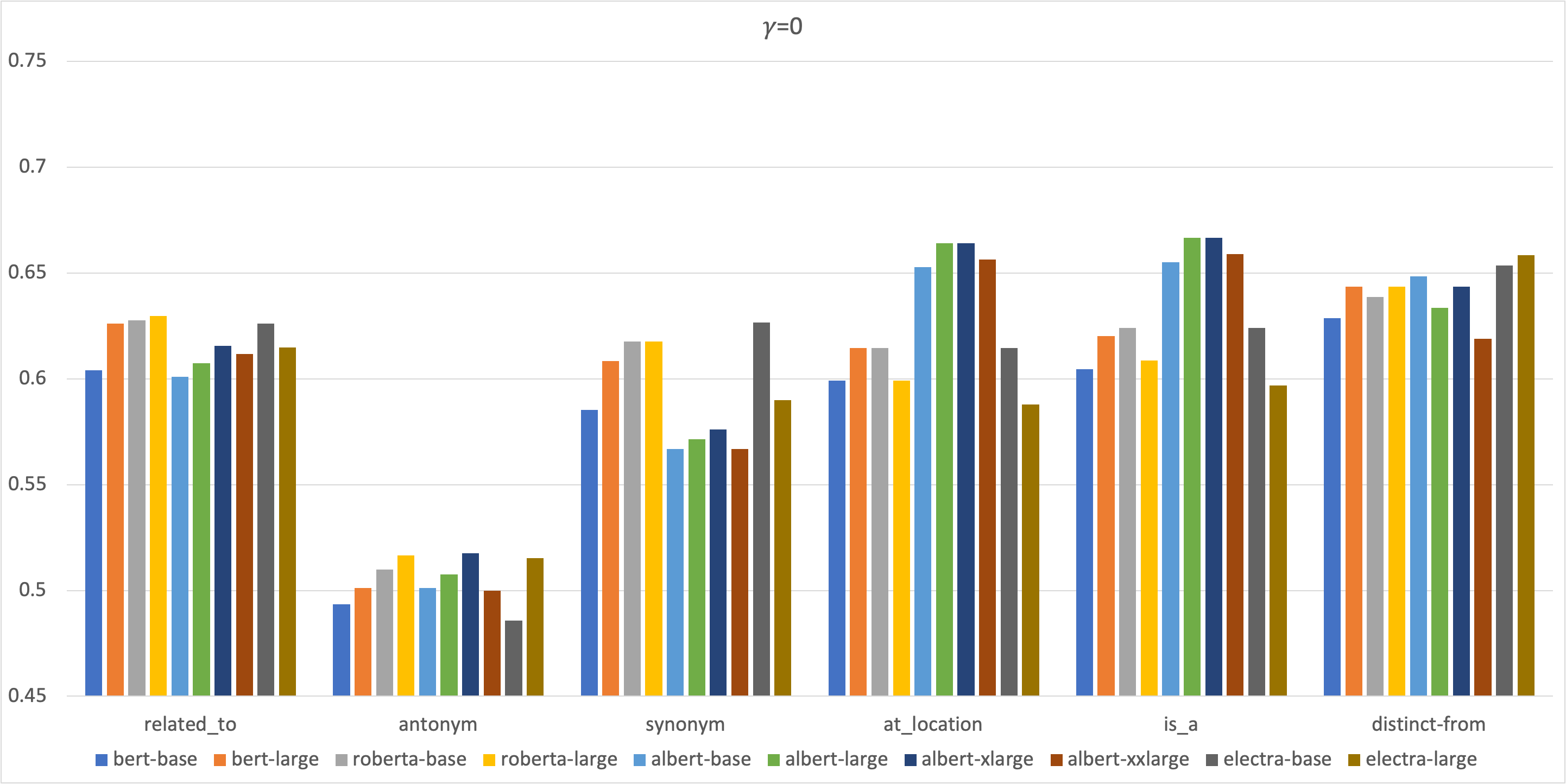}
\end{minipage}
\hfill

\begin{minipage}[b]{\textwidth}
  \includegraphics[width=\linewidth]{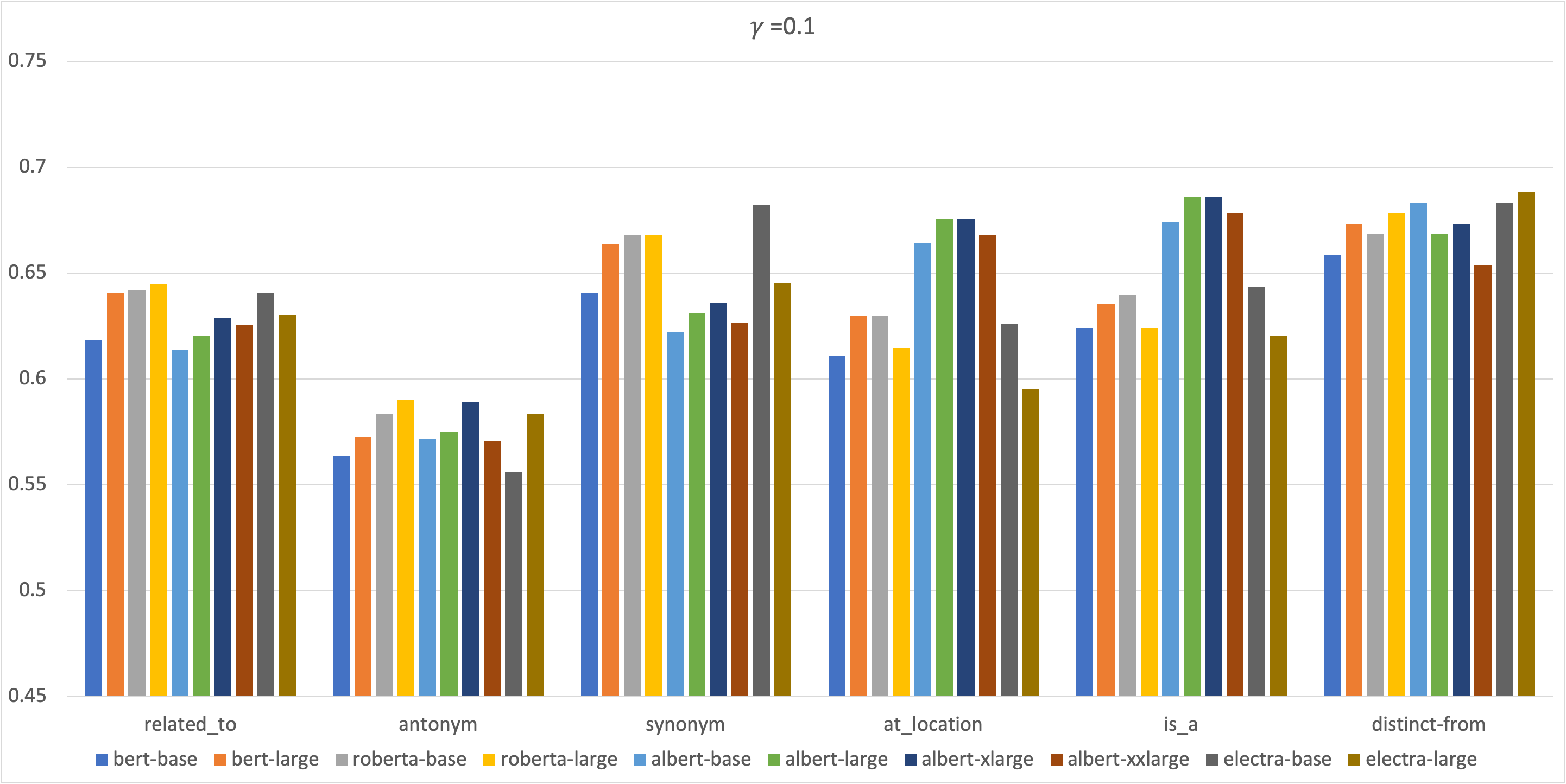}
\end{minipage}
\caption{Directional Accuracy of Contextual Embeddings under different $\gamma$ values (0 and 0.1)}
\label{fig:dirgamma1}
\end{figure*}

\begin{figure*}[!]
\centering

\begin{minipage}[b]{\textwidth}
  \includegraphics[width=\linewidth]{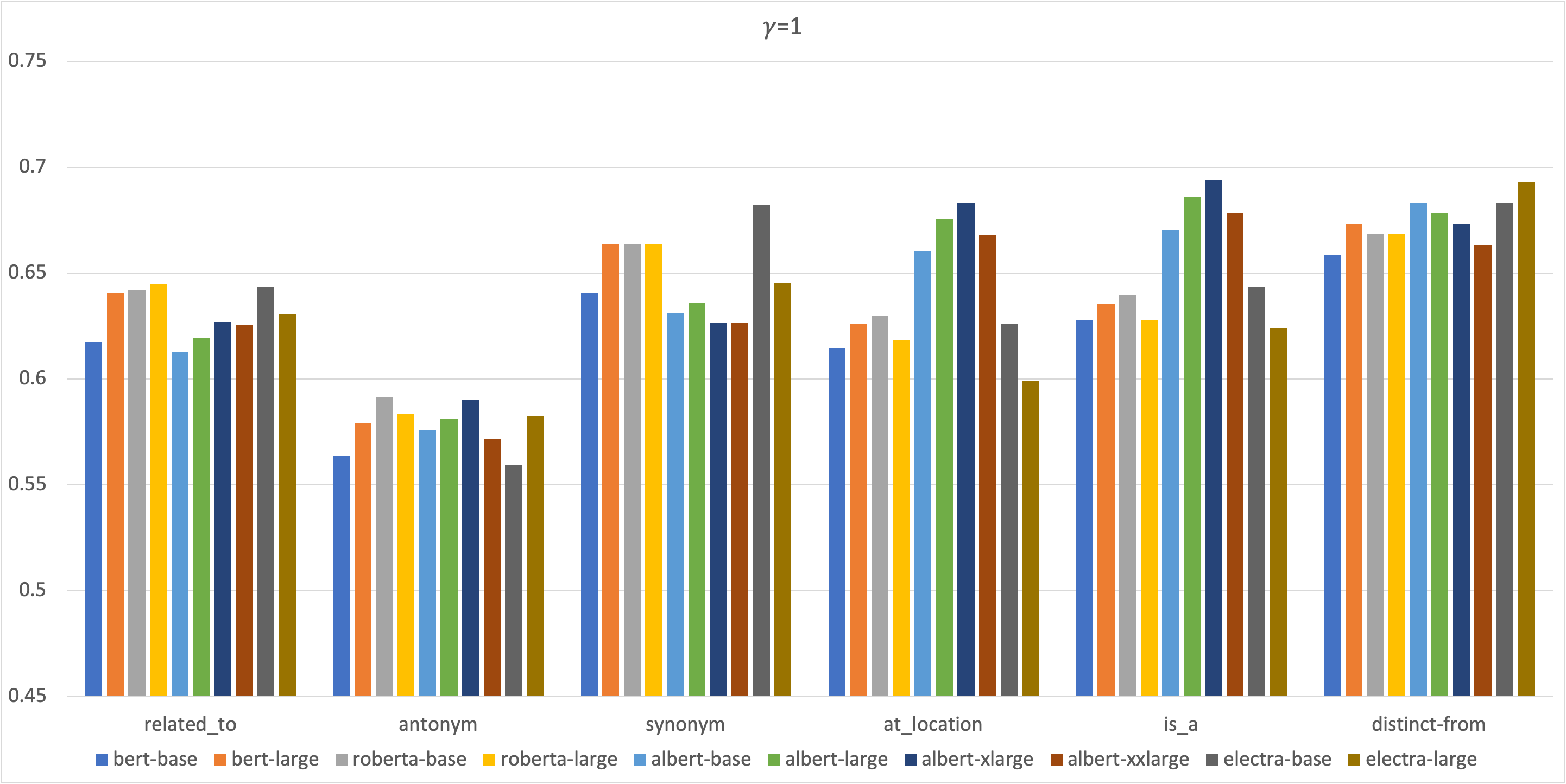}
\end{minipage}
\hfill

\begin{minipage}[b]{\textwidth}
  \includegraphics[width=\linewidth]{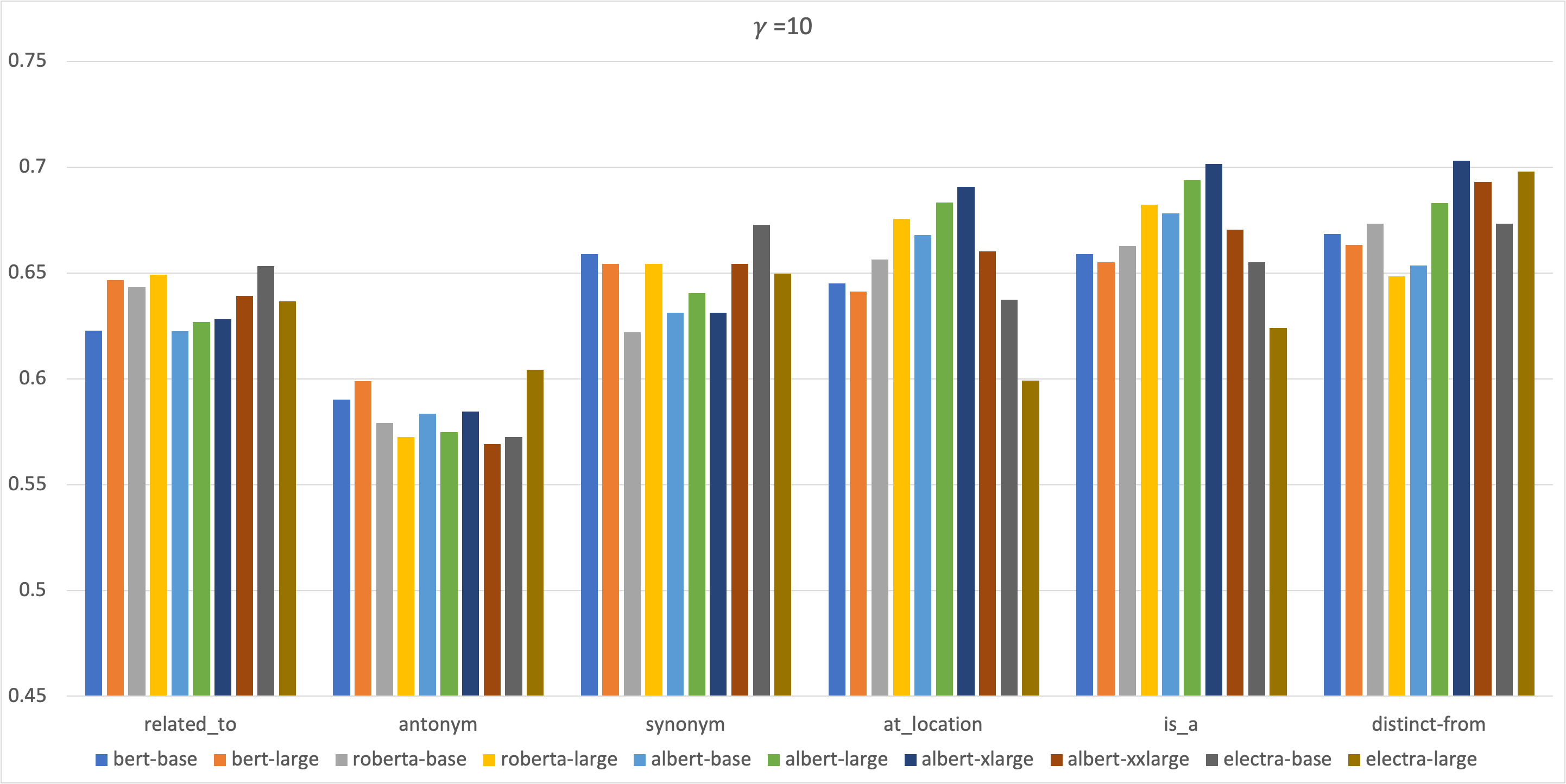}
\end{minipage}
\caption{Directional Accuracy of Contextual Embeddings under different $\gamma$ values (1 and 10)}
\label{fig:dirgamma2}
\end{figure*}

\begin{figure*}[!]
\centering
\begin{minipage}[b]{0.45\textwidth}
    \includegraphics[width=\textwidth]{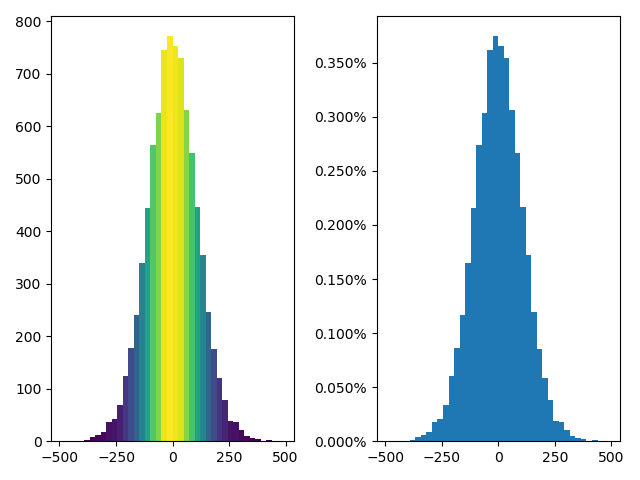}
    \caption*{BERT-base}
  \end{minipage}
  \hfill
  \begin{minipage}[b]{0.45\textwidth}
    \includegraphics[width=\textwidth]{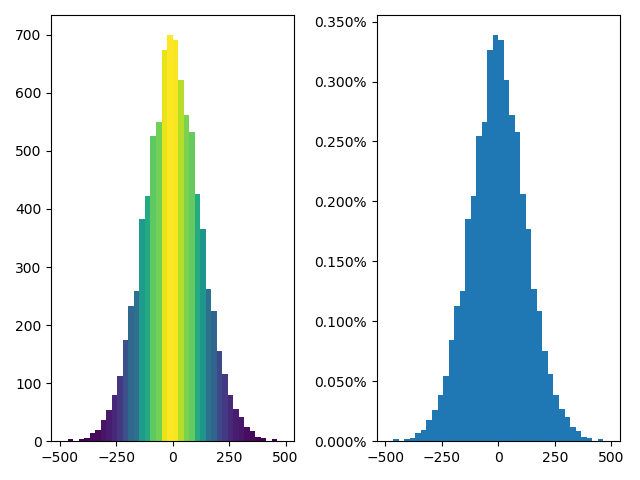}
    \caption*{BERT-large}
  \end{minipage}
 
\caption{BERT prediction of LAR. x=LAR value, y=LAR value counts (left) or Proportions (right). \# bin is 40}
\label{fig:larbert}
\end{figure*}

\begin{figure*}[!ht]
\centering
\begin{minipage}[b]{0.45\textwidth}
    \includegraphics[width=\textwidth]{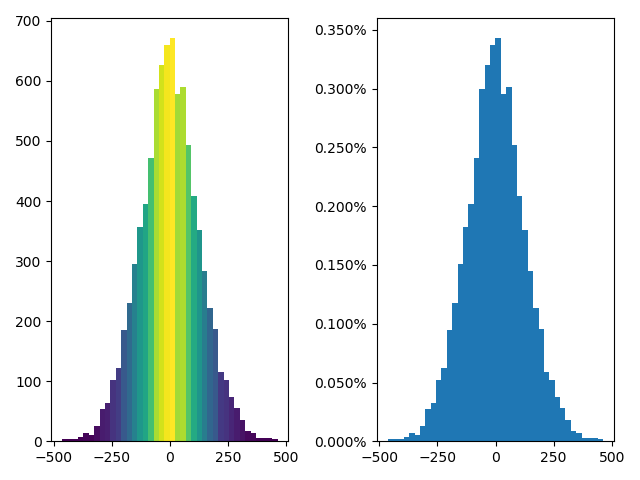}
    \caption*{roBERTa-base}
  \end{minipage}
  \hfill
  \begin{minipage}[b]{0.45\textwidth}
    \includegraphics[width=\textwidth]{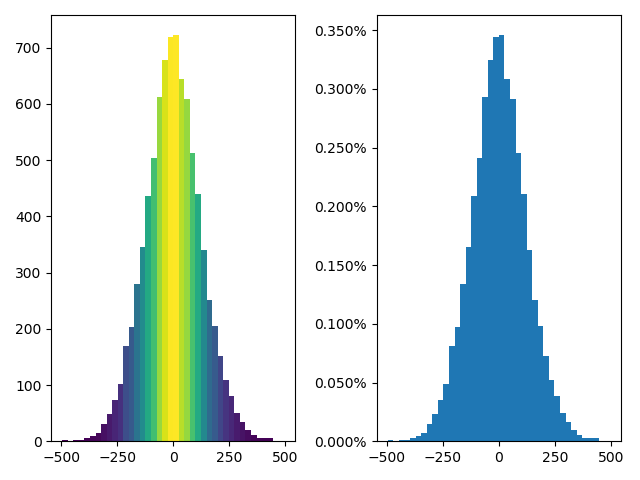}
    \caption*{roBERTa-large}
  \end{minipage}

 \caption{roBERTa prediction of LAR. x=LAR value, y=LAR value counts (left) or Proportions (right). \# bin is 40}
 \label{fig:larroberta}

\end{figure*}

\begin{figure*}[!ht]
\centering
\begin{minipage}[b]{0.45\textwidth}
    \includegraphics[width=\textwidth]{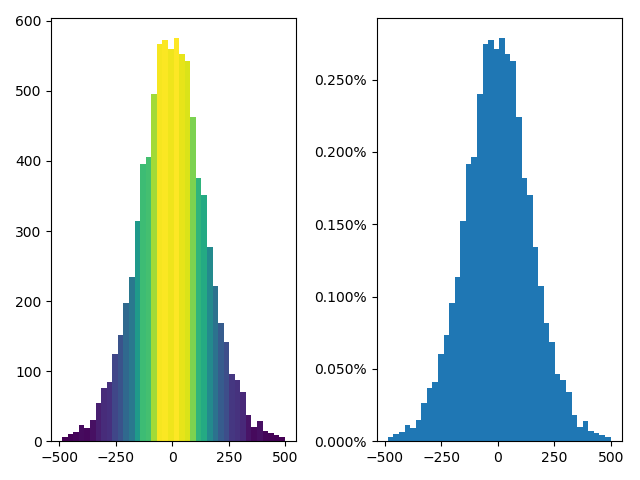}
    \caption*{ALBERT-base}
  \end{minipage}
  \hfill
  \begin{minipage}[b]{0.45\textwidth}
    \includegraphics[width=\textwidth]{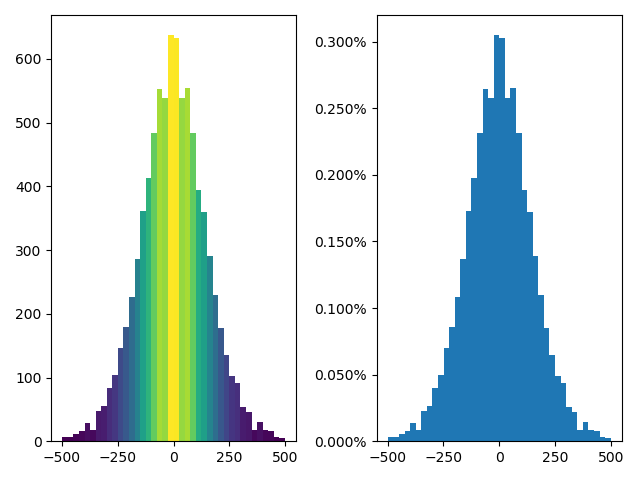}
    \caption*{ALBERT-large}
  \end{minipage}
\hfill
\begin{minipage}[b]{0.45\textwidth}
    \includegraphics[width=\textwidth]{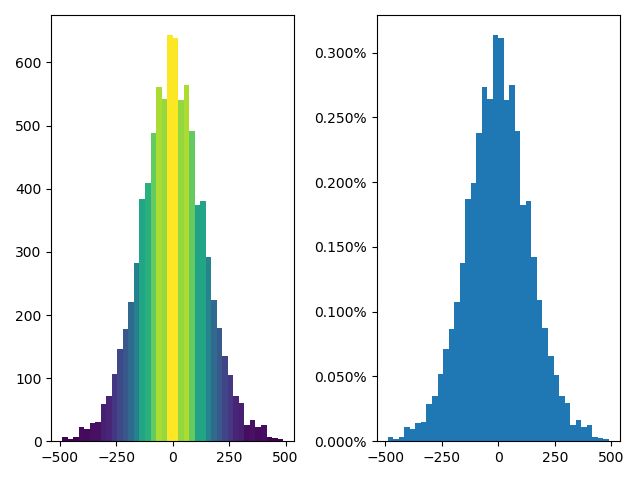}
    \caption*{ALBERT-xlarge}
  \end{minipage}
  \hfill
  \begin{minipage}[b]{0.45\textwidth}
    \includegraphics[width=\textwidth]{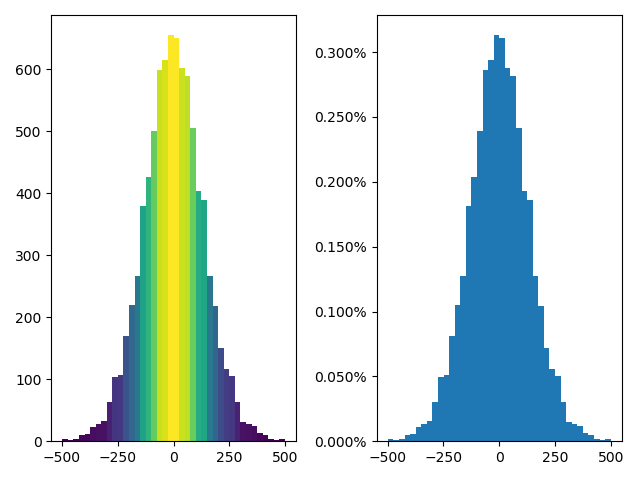}
    \caption*{ALBERT-xxlarge}
  \end{minipage}
\caption{ALBERT prediction of LAR. x=LAR value, y=LAR value counts (left) or Proportions (right). \# bin is 40}
\label{fig:laralbert}
\end{figure*}

\begin{figure*}[!ht]
\centering
\begin{minipage}[b]{0.45\textwidth}
    \includegraphics[width=\textwidth]{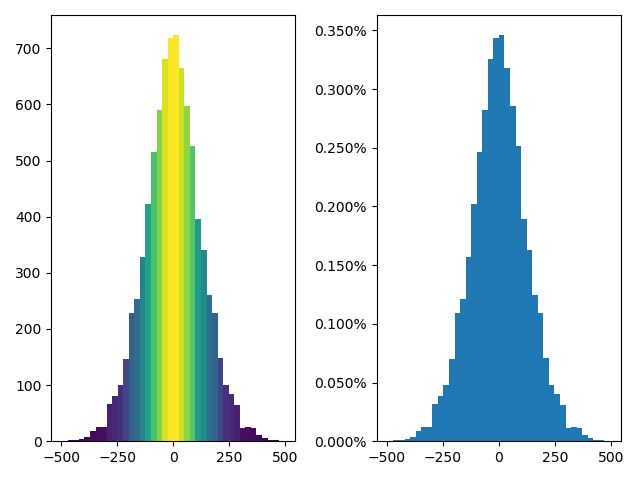}
    \caption*{ELECTRA-base}
  \end{minipage}
  \hfill
  \begin{minipage}[b]{0.45\textwidth}
    \includegraphics[width=\textwidth]{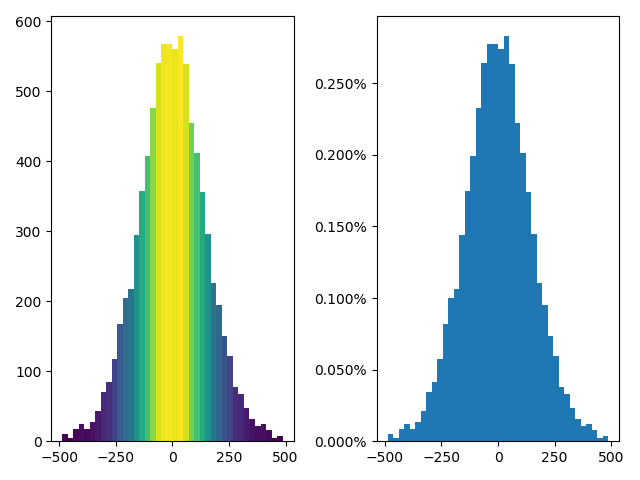}
    \caption*{ELECTRA-large}
  \end{minipage}
\caption{ELECTRA prediction of LAR. x=LAR value, y=LAR value counts (left) or Proportions (right). \# bin is 40}
\label{fig:larelectra}

\end{figure*}

We further discretized LAR into direction indicators \{$-1,0,1$\} and define the direction accuracy $\mbox{Dir}_{D,E}(\mathcal{S})$ of pair set $\mathcal{S}$ for evocation data $D$ and embedding $E$ as 
\begin{equation}
    \mbox{Dir}(\mathcal{S})=\mathop{\mathbf{E}}_{(a,b) \in \mathcal{S}} [\mathbf{I}(Dir_D(a;b)=Dir_E(a;b))]
\end{equation}
,where
\begin{equation}
\label{eq:diracc}
    Dir_X(a;b)= 
\begin{cases}
    1, &\mbox{if } LAR(a;b)> \gamma\\
    -1, &\mbox{if } LAR(a;b)< -\gamma\\
    0,   &            otherwise
\end{cases}
\end{equation}
where we choose $\gamma \in \{0,0.1,1,10\}$ by observation in Table~\ref{tab:1}. If the sign of embedding LAR is the same as the data, we say the prediction is correct, wrong otherwise. From Fig.~\ref{fig:dir_acc}, we show \texttt{BERT-base} and GLoVe are competitive on Directional Accuracy in general. We show the directional accuracy of contextual embeddings in Fig. \ref{fig:dirgamma1} and \ref{fig:dirgamma2}. We also show the LAR distribution from the ground truth and Contextual Embedding's predictions in Fig. \ref{fig:larbert}, \ref{fig:larroberta}, \ref{fig:laralbert}, \ref{fig:larelectra}. 

\begin{figure*}[!]
\centering
  \includegraphics[width=0.5\linewidth]{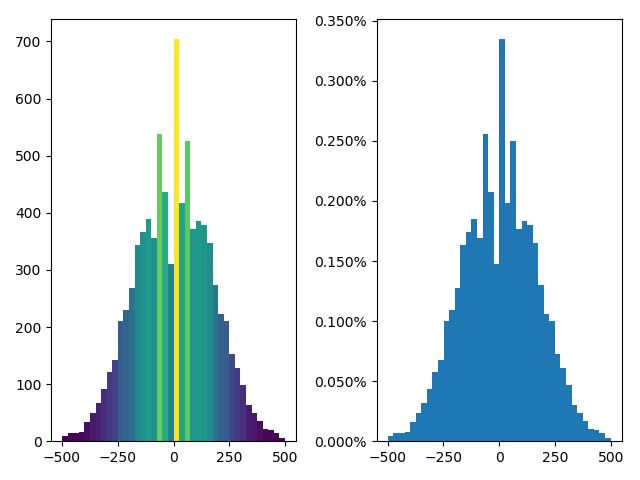}
  \caption{SWOW Ground-truth LAR distribution. x=LAR value, y=LAR value counts (left) or Proportions (right). \# bin is 40}
  \label{fig:largt}
\end{figure*}

In general, from Fig. \ref{fig:dirgamma1} and \ref{fig:dirgamma2} we see that the relation of ``antonym'' is generally harder than others, under all $\gamma$ settings; and the LAR accuracy all relations get higher when $\gamma$ is set to larger values, because, according to the distribution of \ref{fig:largt} and predictions such as \ref{fig:larbert} we see relations around zero dominates, and the larger $\gamma$ is, the more values around zero are incorporated.

\begin{figure*}[!]
\centering
\begin{minipage}[b]{0.45\textwidth}
    \includegraphics[width=\textwidth]{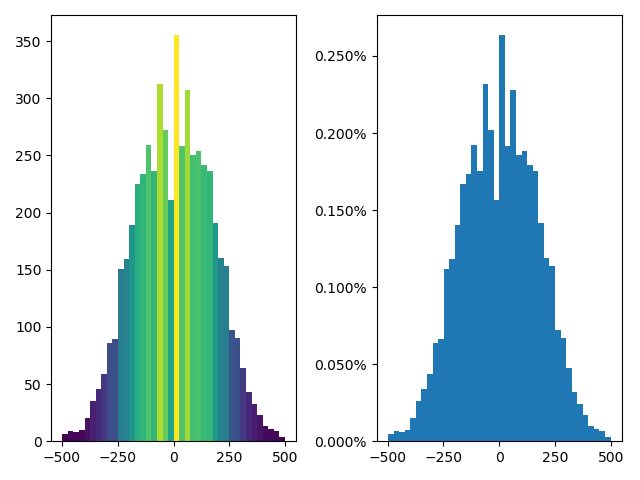}
    \caption*{SWOW}
  \end{minipage}
  \hfill
  \begin{minipage}[b]{0.45\textwidth}
    \includegraphics[width=\textwidth]{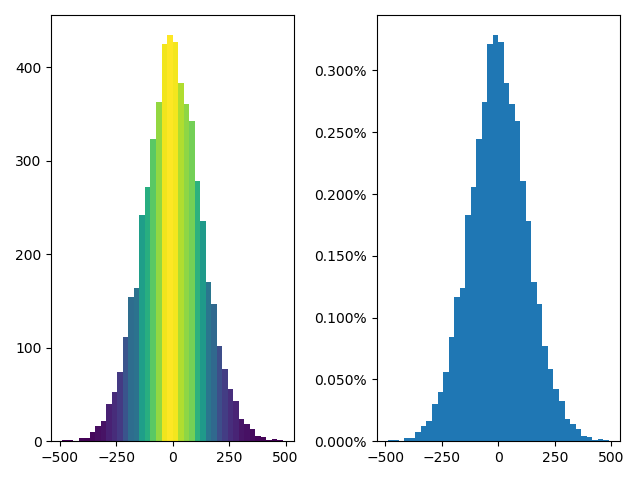}
    \caption*{\texttt{BERT-large}}
  \end{minipage}
\caption{LAR distribution comparison on the pairs annotated with ``relatedTo'' in Table \ref{tab:compare_cxt_emb} }
\label{fig:compare_lar_relatedto}
\end{figure*}

\begin{figure*}[!]
\centering
\begin{minipage}[b]{0.45\textwidth}
    \includegraphics[width=\textwidth]{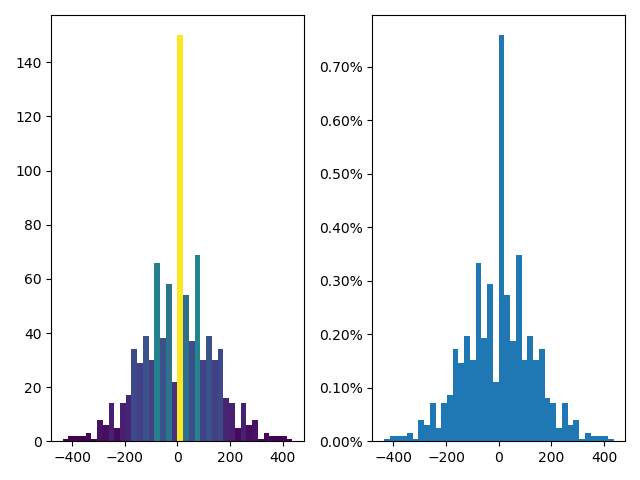}
    \caption*{SWOW}
  \end{minipage}
  \hfill
  \begin{minipage}[b]{0.45\textwidth}
    \includegraphics[width=\textwidth]{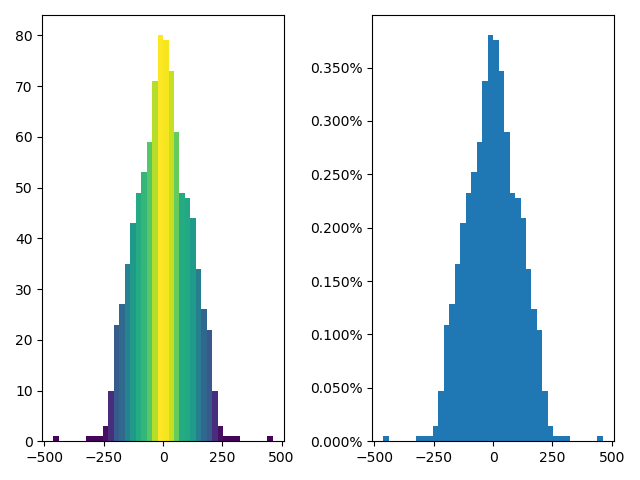}
    \caption*{\texttt{BERT-large}}
  \end{minipage}
\caption{LAR distribution comparison on the pairs annotated with ``antonym'' in Table \ref{tab:compare_cxt_emb} }
\label{fig:compare_lar_antonym}
\end{figure*}

\begin{figure*}[!]
\centering
\begin{minipage}[b]{0.45\textwidth}
    \includegraphics[width=\textwidth]{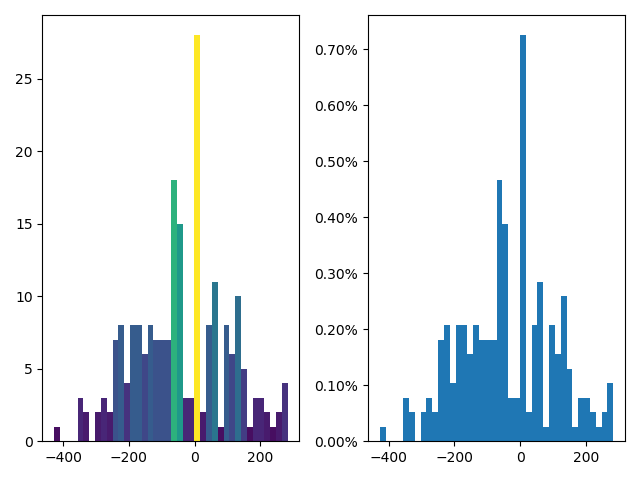}
    \caption*{SWOW}
  \end{minipage}
  \hfill
  \begin{minipage}[b]{0.45\textwidth}
    \includegraphics[width=\textwidth]{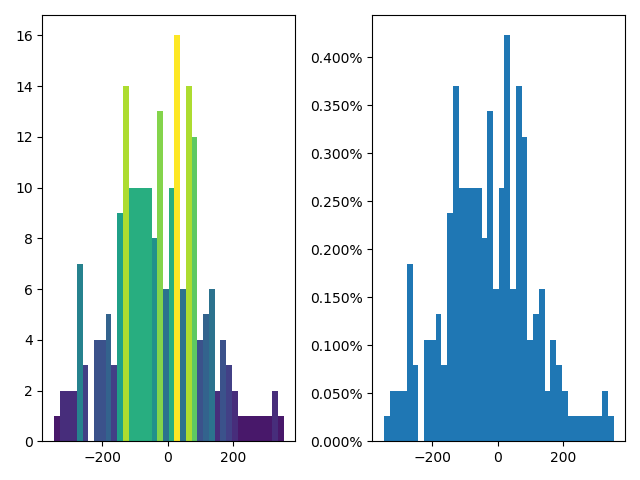}
    \caption*{\texttt{BERT-large}}
  \end{minipage}
\caption{LAR distribution comparison on the pairs annotated with ``synonym'' in Table \ref{tab:compare_cxt_emb} }
\label{fig:compare_lar_synonym}
\end{figure*}

\begin{figure*}[!]
\centering
\begin{minipage}[b]{0.45\textwidth}
    \includegraphics[width=\textwidth]{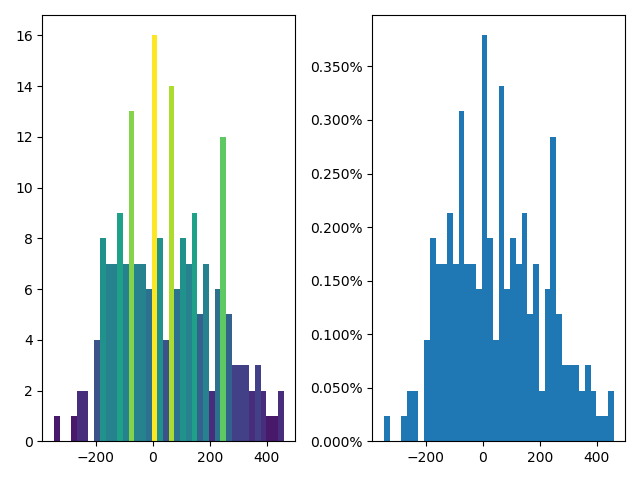}
    \caption*{SWOW}
  \end{minipage}
  \hfill
  \begin{minipage}[b]{0.45\textwidth}
    \includegraphics[width=\textwidth]{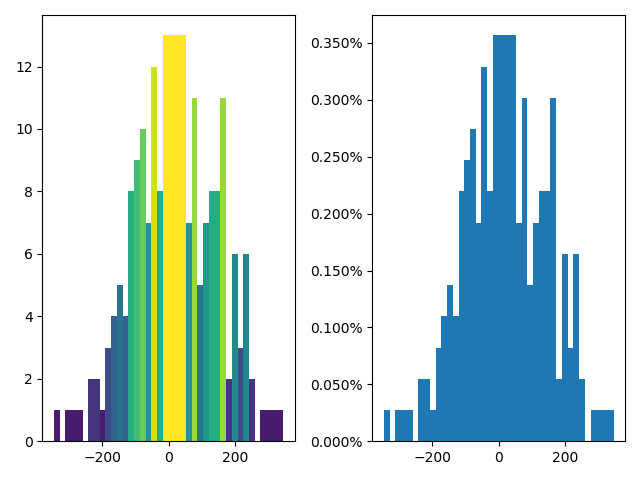}
    \caption*{\texttt{BERT-large}}
  \end{minipage}
\caption{LAR distribution comparison on the pairs annotated with ``isA'' in Table \ref{tab:compare_cxt_emb} }
\label{fig:compare_lar_isa}
\end{figure*}

\begin{figure*}[!]
\centering
\begin{minipage}[b]{0.45\textwidth}
    \includegraphics[width=\textwidth]{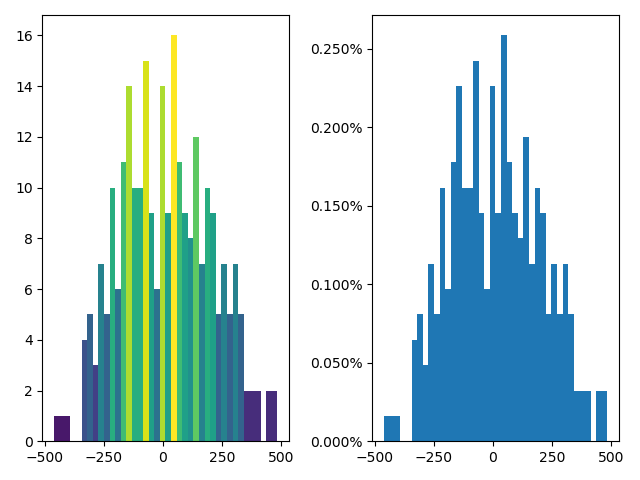}
    \caption*{SWOW}
  \end{minipage}
  \hfill
  \begin{minipage}[b]{0.45\textwidth}
    \includegraphics[width=\textwidth]{rel_atlocation_gtSWOW_bert_large.png}
    \caption*{\texttt{BERT-large}}
  \end{minipage}
\caption{LAR distribution comparison on the pairs annotated with ``atLocation'' in Table \ref{tab:compare_cxt_emb} }
\label{fig:compare_lar_atlocation}
\end{figure*}

\begin{figure*}[!]
\centering
\begin{minipage}[b]{0.45\textwidth}
    \includegraphics[width=\textwidth]{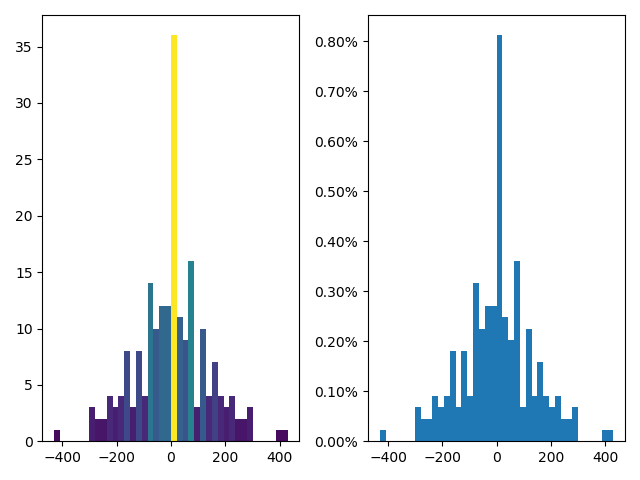}
    \caption*{SWOW}
  \end{minipage}
  \hfill
  \begin{minipage}[b]{0.45\textwidth}
    \includegraphics[width=\textwidth]{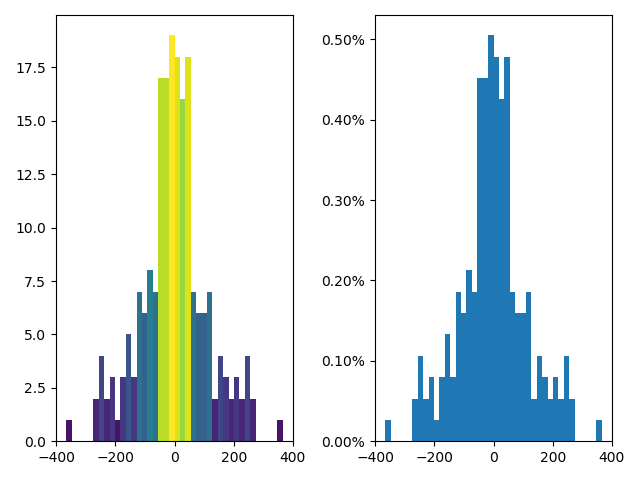}
    \caption*{\texttt{BERT-large}}
  \end{minipage}
\caption{LAR distribution comparison on the pairs annotated with ``distinctFrom'' in Table \ref{tab:compare_cxt_emb} }
\label{fig:compare_lar_distinctFrom}
\end{figure*}

We also showcase the distribution of LAR values for 6 relations ``relatedto'', ``antonym'', ``synonym'', ``isa'', ``atlocation'' and ``distinctfrom'' for the SWOW pairs used in Table \ref{tab:compare_cxt_emb} and \texttt{BERT-large} language model in Fig. \ref{fig:compare_lar_relatedto}, \ref{fig:compare_lar_antonym}, \ref{fig:compare_lar_synonym}, \ref{fig:compare_lar_isa}, \ref{fig:compare_lar_atlocation}, \ref{fig:compare_lar_distinctFrom}. From those figures we see that for the pairs annotated with ``isA'' and ``synonym'' the LAR distribution is slightly skewed to negative side, whereas ``atLocation'' is to the positive side; symmetric relations such as ``distinctFrom'', ``antonym'' are symmetric on pair distribution as well; For ``relatedTo'' which is the most uncertain relation, LAR distribution also is symmetric-like. Those figures provides an intuitive feeling of what the evocation data and BERT predictions are like.

\section{The Two Factors for word relatedness estimation with BERT: Further Analysis}

It is interesting that when we look at Fig. \ref{fig:da_stat} that the upper three graphs indicate a counter-intuitive phenomenon: using more contexts does not improve the direction accuracy further, especially when the number of contexts is higher than 5000. 

\begin{figure*}[!]
  \includegraphics[width=\linewidth]{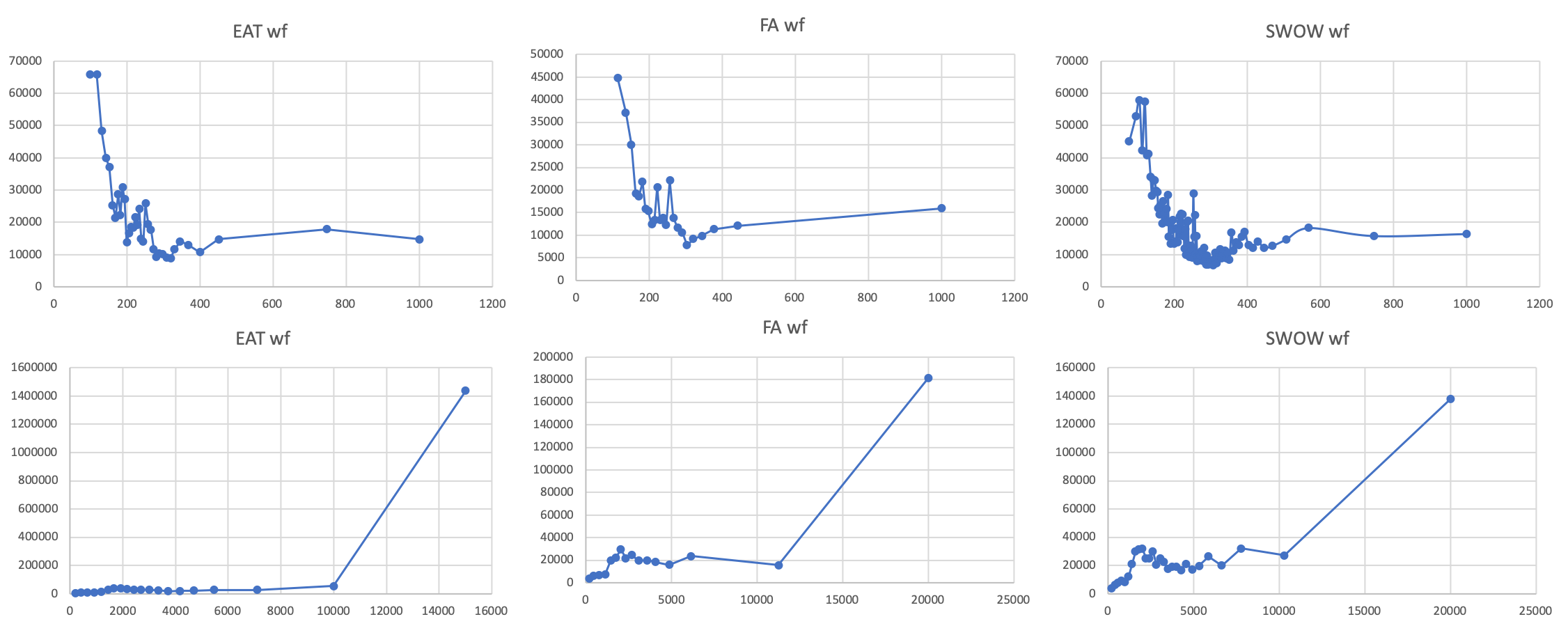}
  \caption{Word frequency (y-axis) v.s. word character distance in sentence (x-axis upper) and  number of contexts (x-axis lower) on EAT, FA and SWOW datasets. \textit{Are average word frequency and Pair Distance correlated?} The correlation between the average character level word pair distance in each size-200 bin and the average word frequency for the pair in the bin. The figure suggests that in general, the closer the words are in the context, the more frequent the words are in the corpus, which in turn suggests that frequent words tend to be close to other frequent words in word evocation data, a secondary but interesting result. Also, in this figure, we have clearly shown the correlation between word average frequency and the number of contexts from three lower sub-graphs. It shows that when the number of contexts increases, the word frequency increases sharply. In general, for the word pairs that have more number of contexts, the average word frequency for the words in those pairs is the higher as well, which provides an indirect evidence for our assumption that word frequency plays a role in relatedness estimation of BERT. There may be ways we can explore to reduce such a ``local context effect'', such as filtering the context that may be less likely to correlate the words in a pair. Although relevant, the discussion on mitigating the bias can not be trivially addressed in this paper, and we may pursue it in future work.}
  \label{fig:stat_wf}
\end{figure*}

The context ``quality'' may be to blame, especially the relation of word pair to the context. For example, for frequent words, the probability of a word in a pair may rely much more on the words around it instead of the other word. This phenomenon may introduce a systematic bias when all contexts containing that word show this ``local-prediction characteristic''. Thus, a good indicator of this phenomenon is the correlation between term frequency of a word in pair and the number of contexts for the pair. The more frequent the word is, the more common the word is, and in turn the stronger it may relate to its local context instead of the other word. Details are shown in Figure \ref{fig:stat_wf}

\section{Long-tailness of LAR}
\paragraph{Is LAR robust to the long-tail issue?} It is essential to note that, although LAR is produced from count-based probabilities, it may bypass the side-effect of the long-tail distribution issue that a $P(b|a)$ suffers from: the small but uniform distribution over random variable $b$ corrupts the rank-based evaluations. 

In detail, the median of the entire SWOW dataset of $P(b|a)$ is 1, meaning most of the pairs are singletons. 

In this work, the problem is alleviated because 1) the set of pairs we evaluate are a subset of pairs where they co-occur in context. Thus, the likelihood of words being related has increased, resulting in the count(cue=$a$ , target=$b$) has mean 11.17, median 5 and count(cue=$b$, target=$a$) has mean 10.83 and median 5, both being far away from 1. This shows that the long-tailed issue can be addressed by ''context filtering''; 2) the long-tailed issue can also be addressed using two conditional likelihoods, which decreased the likelihood of long-tailedness. This can be seen by answering the question, ``How many pairs do we have where both previous counts are long-tail likelihoods?'' Thus, we examining if those two counts both have count 1. If yes, the pair is identified as a ``rare pair'', no otherwise. We found that 250 out of 7640 pairs (3\%) contain pairs that both have 1 count, which justifies the dataset of 7640 pairs being a valid source for asymmetry judgment.

\section{Pseudo Code for CAM calculation}

In Alg. \ref{algo:cam} we give a pseduo code to calculated Spearman Correlation on Asymmetry Measure between two resources $\mathcal{E}_i$ and $\mathcal{E}_j$ on a set of word pairs $\mathcal{S}$. The algorithm is universal for all pair of resources. The critical part is the choice of $\mathcal{S}$. For relation specific $r$ the set is $\mathcal{S}_r$. As is noted in Eq. \ref{eq:common_pairs}, the choice of this set is based on the resources to be compared: the intersection of vocabularies is used to filter the pairs so that no OOV words affect asymmetry judgment.

\begin{algorithm}[!]
   \caption{Spearman Correlation of Asymmetry Measure (CAM) of Eq. \ref{eq:cam}}
    \begin{algorithmic}
      \Function{CAM}{$\mathcal{S}, \mathcal{E}_i, \mathcal{E}_j $ } \Comment{ $\mathcal{S}$ is the set of word pairs, $\mathcal{E}_i$, $\mathcal{E}_j$ are two resources, either embedding or evocation data}
        \State let $\mathcal{M}_i$, $\mathcal{M}_j$ be new maps 
        \For{ (a,b) in $\mathcal{S}$ }
            \State obtain $P_{\mathcal{E}_i}(b|a)$ and $P_{\mathcal{E}_i}(a|b)$ using Section \ref{sec:pba} 
            \State obtain $P_{\mathcal{E}_j}(b|a)$ and $P_{\mathcal{E}_j}(a|b)$ using Section \ref{sec:pba}
            \State obtain LAR$(a;b)$ from Eq. \ref{eq:alar}
            \State $\mathcal{M}_i[(a;b)]=LAR(a;b)$
            \State $\mathcal{M}_j[(a;b)]=LAR(a;b)$
        \EndFor
        \State Return Spearman($\mathcal{M}_i$,$\mathcal{M}_j$)
       \EndFunction
\end{algorithmic}
\label{algo:cam}
\end{algorithm}

\end{document}